\documentclass[lettersize,journal]{IEEEtran}
\usepackage{amsmath,amsfonts,bm}
\usepackage{algorithmic}
\usepackage{algorithm}
\usepackage{array}
\usepackage{textcomp}
\usepackage{stfloats}
\usepackage{url}
\usepackage{verbatim}
\usepackage{graphicx}
\usepackage{cite}
\usepackage{booktabs}
\usepackage{multirow}
\usepackage{pifont}
\usepackage{etoolbox}
\usepackage{subfigure}
\usepackage[breaklinks=true,colorlinks,bookmarks=false]{hyperref}
\begin{document}
% \captionsetup[figure]{labelsep=colon}
\title{Balanced Classification: A Unified Framework for Long-Tailed Object Detection}

\author{Tianhao Qi, Hongtao Xie*, Pandeng Li, Jiannan Ge and Yongdong Zhang, Senior Member, IEEE
        % <-this % stops a space
\thanks{T. Qi, H. Xie*, P. Li, J. Ge and Y. Zhang are with the School of Information Science and Technology, University of Science and Technology of China, Hefei 230022, China. (email: qth@mail.ustc.edu.cn; htxie@ustc.edu.cn; lpd@mail.ustc.edu.cn; gejn@mail.ustc.edu.cn; zhyd73@ustc.edu.cn)}% <-this % stops a space
\thanks{H. Xie is the corresponding author. (email: htxie@ustc.edu.cn).}}

\markboth{Journal of \LaTeX\ Class Files,~Vol.~14, No.~8, August~2021}%
{Shell \MakeLowercase{\textit{et al.}}: A Sample Article Using IEEEtran.cls for IEEE Journals}


\maketitle

\begin{abstract}
  Conventional detectors suffer from performance degradation when dealing with long-tailed data due to a classification bias towards the majority head categories.
  In this paper, we contend that the learning bias originates from two factors: 1) the unequal competition arising from the imbalanced distribution of foreground categories, and 2) the lack of sample diversity in tail categories.
  To tackle these issues, we introduce a unified framework called \textbf{BA}lanced \textbf{CL}assification (\textbf{BACL}), which enables adaptive rectification of inequalities caused by disparities in category distribution and dynamic intensification of sample diversities in a synchronized manner.
  Specifically, a novel foreground classification balance loss (FCBL) is developed to ameliorate the domination of head categories and shift attention to difficult-to-differentiate categories by introducing pairwise class-aware margins and auto-adjusted weight terms, respectively.
  This loss prevents the over-suppression of tail categories in the context of unequal competition.
  Moreover, we propose a dynamic feature hallucination module (FHM), which enhances the representation of tail categories in the feature space by synthesizing hallucinated samples to introduce additional data variances.
  In this divide-and-conquer approach, BACL sets a new state-of-the-art on the challenging LVIS benchmark with a decoupled training pipeline, surpassing vanilla Faster R-CNN with ResNet-50-FPN by 5.8\% AP and 16.1\% AP for overall and tail categories.
  Extensive experiments demonstrate that BACL consistently achieves performance improvements across various datasets with different backbones and architectures.
  Code and models are available at \url{https://github.com/Tianhao-Qi/BACL}.
\end{abstract}


\section{Introduction}

\begin{figure}[!t]
  \centering
  \includegraphics[width=\columnwidth]{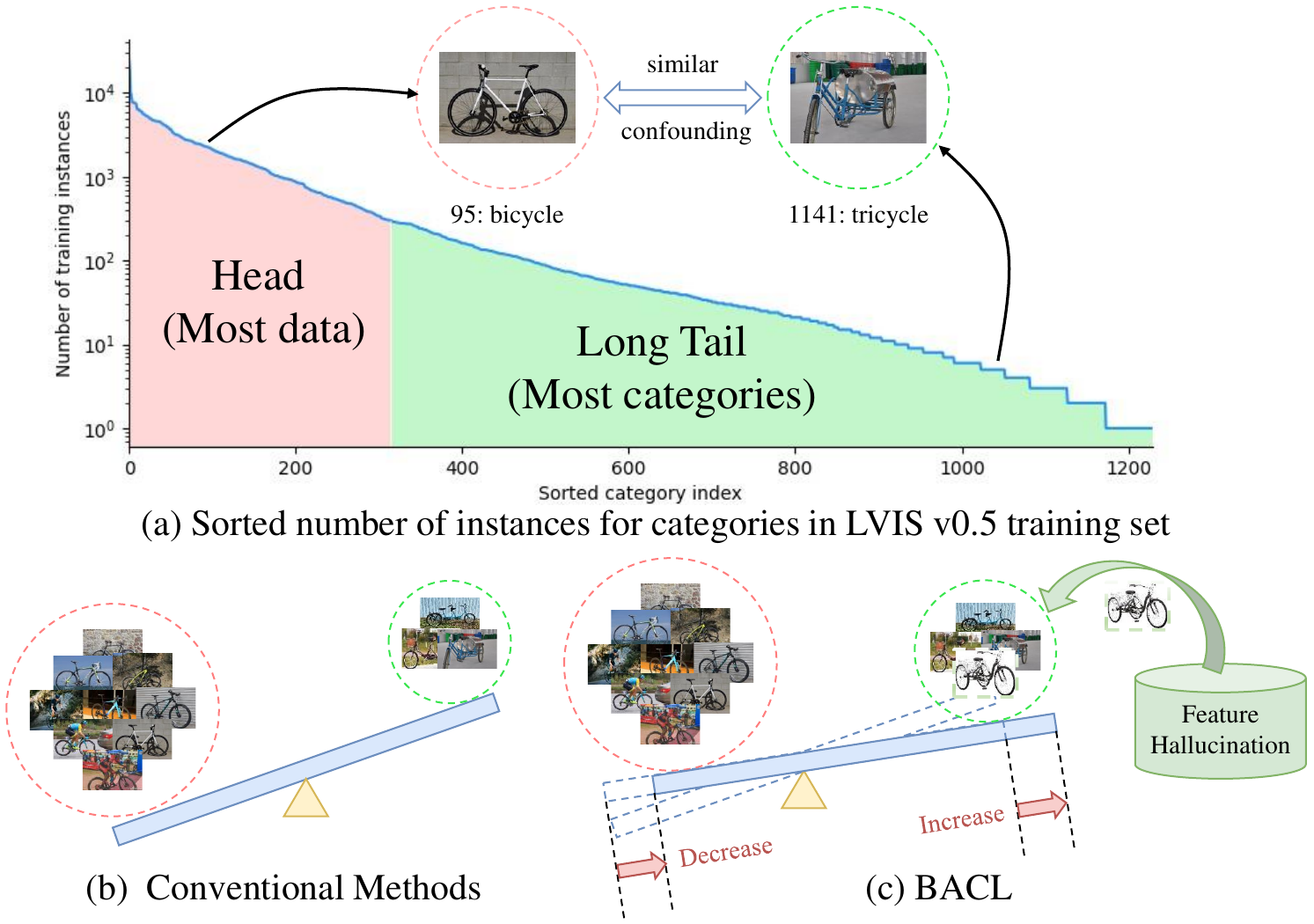}
  \caption{LVIS contains visually similar categories distributed in both the head and long tail ({\em bicycle} vs. {\em tricycle}). Due to their limited discriminatory power, conventional methods often misclassify instances of {\em tricycles} as {\em bicycles}. BACL focuses on these confounding categories, adaptively lifting their contributions to the loss formula. Meanwhile, BACL synthesizes diverse features to enhance the representation of tail categories, thereby assisting the detector in achieving a more balanced state across all categories.}
  \label{pic:1}
\end{figure}

\IEEEPARstart{R}{ecently}, deep neural networks~\cite{yang2020captionnet,qiu2017learning,zhan2019unmanned,wu2021deep,li2021online,wang2021data,li2022deep,ge2021semantic,cai2019exploring,li2016deepsaliency} have shown promising results on well-balanced object detection benchmarks, such as Pascal VOC~\cite{everingham2010pascal} and MS COCO~\cite{lin2014microsoft}.
However, as opposed to these artificially balanced datasets, real-world data exhibits an inherent long-tailed distribution.
That is to say, a few categories (head categories) occupy the vast majority of samples, while the remaining many categories (tail categories) only correspond to a small proportion of samples, as illustrated in Fig. \ref{pic:1}(a). 
Such a skewed distribution inevitably poses an immense challenge for data-hungry deep neural networks.
Unfortunately, current deep learning-based detectors ({\em e.g.}, Faster R-CNN~\cite{ren2015faster}) suffer from performance degradation especially on under-represented tail categories, when confronted with long-tail datasets like LVIS~\cite{gupta2019lvis}, which better resembles real-world scenarios.
This undesirable performance degradation significantly affects the practicality of deep learning-based detectors, consequently impeding the progress of related technologies.
Therefore, exploring efficacious training strategies to cope with long-tailed object detection has attracted broad research interests.

To facilitate boosting performance in realistic scenarios, pioneering works~\cite{wang2020devil, li2020overcoming} extensively investigate the detrimental impact of  the long-tailed distribution in training data on various components of a two-stage detector, namely, Faster R-CNN.
These studies reach a consensus that the classifier, which is severely biased towards head categories, serves as the primary bottleneck influencing detection metrics.
Consequently, mitigating the classification bias has emerged as a prominent trend in long-tailed object detection.

For this purpose, an intuitive solution is to re-balance the training distribution.
Some works achieve this goal by image-level~\cite{gupta2019lvis} or object-level~\cite{chang2021image, feng2021exploring} re-sampling.
Another line of research explores specialized margin designs~\cite{cao2019learning, ren2020balanced, feng2021exploring} or re-weighting strategies~\cite{tan2020equalization, wang2021adaptive, tan2021equalization, zhang2021distribution, wang2021seesaw} for the classification loss, aiming to simulate a more balanced training set.
In addition to data re-balancing, several novel detector architectures ~\cite{li2020overcoming,wu2020forest} have been proposed to incorporate valuable prior knowledge into the training process.
Orthogonal to these methods, Kang {\em et al.}~\cite{kang2019decoupling} creatively decouple the learning process into the representation learning and classifier learning stages for long-tailed image classification, dramatically promoting accuracy.
This two-stage decoupled training pipeline has also been proven effective in long-tailed object detection and applied to subsequent researches~\cite{li2020overcoming, wang2021adaptive, feng2021exploring}.

However, the majority of existing methods primarily focus on mitigating the unequal competition among imbalanced foreground categories, overlooking the underlying issue of under-representation in tail categories.	
In essence, they neglect to enhance the diversity of training instances from tail categories.
Thus, this lack of sample variations undermines the discriminatory capability of the classifier for tail categories and limits further improvements in detection performance.

To further enhance performance, we adopt a divide-and-conquer approach to take both issues into consideration:
\begin{itemize}
    \item The unequal competition among foreground categories arising from extremely imbalanced frequencies;
    \item The latent issue of under-representation in tail categories, characterized by a scarcity of diverse visual instances.
\end{itemize}
To address them, we introduce a unified framework called \textbf{BA}lanced \textbf{CL}assification (\textbf{BACL}), as depicted in Fig. \ref{pic:3}, on the basis of the aforementioned decoupled learning scheme~\cite{kang2019decoupling}.
For the convenience of debiasing, BACL utilizes a pair of long-term and short-term indicators to monitor the learning status of the classifier from two distinct perspectives: the inclination towards classifying foreground categories and the correctness of classification for each category.

Based on this pair of indicators, we integrate two key components into the proposed framework.
Firstly, we introduce a novel foreground classification balance loss (FCBL) to perceive discrepancies among foreground categories and adaptively ameliorate the domination of the head categories over the tail ones through pairwise class-aware margins.
FCBL also assigns an auto-adjusted weight term to each non-ground-truth category, focusing on the difficult-to-differentiate categories.
This weighting strategy facilitates bias calibration, particularly considering the learning difficulty posed by large vocabulary datasets. 
Furthermore, we propose a dynamic feature hallucination module (FHM) to enhance the representation of tail categories in the feature space. 
This module accurately captures the feature distributions and selectively chooses categories with a bias towards tail categories to synthesize hallucinated features using the reparametrization trick~\cite{kingma2013auto}, which introduces additional data variances.
Through the synergy of these two components, BACL works collaboratively to eliminate the classification bias by rectifying inequalities arising from category distribution disparities and intensifying sample diversities, as illustrated in Fig. \ref{pic:1}(c).
% Thus, more varied training samples must be provided for this purpose, and FHM possesses this capability.
% In particular, this module not only captures the feature distribution for each category online, but also judiciously samples the categories and synthesizes multiple hallucinated features for each sampled category by the reparametrization trick\cite{kingma2013auto}.
% Furthermore, to guarantee that the tail categories acquire more hallucinated features, this module assigns higher probabilities which are inverse to the long-term indicators to them.

% \begin{figure}[t]
%     \centering
%     \includegraphics[width=\columnwidth]{pictures/performance.pdf}
%     \caption{Performance comparison on LVIS v0.5 \texttt{val} set. BACL outperforms the baseline~\cite{ren2015faster} and previous arts, including EQL~\cite{tan2020equalization}, EQLv2~\cite{tan2021equalization}, LOCE~\cite{feng2021exploring}, and Seesaw~\cite{wang2021seesaw}.}
%     \label{pic:2}
% \end{figure}

Extensive experiments on long-tailed object detection and instance segmentation demonstrate the effectiveness of BACL.
On the challenging LVIS v0.5 dataset, BACL outperforms the baseline model by a large margin, which increases the overall accuracy from 22.0\% to 27.8\% with a 5.8\% AP improvement.
The improvement is primarily observed in rare categories (+16.1\% AP) and common categories (+7.0\% AP).
BACL even slightly improves the accuracy for frequent categories by 0.2\% AP.
More convincingly, it surpasses all existing state-of-the-art methods by over 1.0\% AP.

Our main contributions can be summarized as follows:
\begin{itemize}
    \item We introduce a pair of complementary long-term and short-term indicators to monitor the learning status of the classifier from two perspectives, which can serve as an accurate guide for real-time bias calibration.
    \item We propose a two-stage decoupled training framework called BACL for long-tailed object detection, which incorporates a novel foreground classification balance loss to mitigate the unequal competition between foreground categories and a dynamic feature hallucination module to intensify the sample diversity, especially in tail categories.
    \item Extensive experiments with various settings demonstrate the superiority of BACL in comparison to the baseline and previous state-of-the-art methods, as well as its excellent generalization ability.
\end{itemize}
% 1) We introduce a pair of complementary long-term and short-term indicators to monitor the learning status of the classifier from two perspectives;
% 2) we propose a unified framework, including a novel loss function to eliminate the inconsistency of training and testing, and a dynamic feature hallucination module to increase sample diversity;
% 3) extensive experiments show the superiority and generalization ability of our proposed framework, which surpasses the vanilla Faster R-CNN with ResNet-50-FPN by 4.9 AP and 16.6 AP for overall and tail categories.

\section{Related Works}

The remarkable fitting capability of deep learning models~\cite{liu2022fine,li2022neighborhood,yao2018exploring}  has greatly contributed to the advancement of the object detection community.
However, as research in this field continues to expand, the long-tail problem encountered by detectors in real-world applications is gaining more attention.
% In this section, we first detail the development of deep learning based detectors, then elaborate all approaches to the prevalent long-tailed object detection task.

\subsection{General Object Detection}
Based on the usage of predefined anchors, existing deep learning-based object detectors can be categorized into two groups: anchor-based and anchor-free.

Anchor-based detectors encompass both two-stage and one-stage methods, employing predefined anchors of various scales and positions as proposals for classification and regression.
The emergence of the R-CNN family~\cite{girshick2014rich,girshick2015fast,ren2015faster,cai2018cascade} has solidified the mainstream status of two-stage detectors.
They first generate coarse bounding box proposals on the basis of predefined anchors by a region proposal network (RPN), while a proposal-wise prediction head (Fast R-CNN) handles the tasks of classification and location refinement.
Follow-up works aim to further enhance the performance by fusing multi-scale features~\cite{lin2017feature}, incorporating contextual information~\cite{bell2016inside}, utilizing improved training strategies~\cite{shrivastava2016training,dai2017deformable,zhu2019deformable}, and applying attention mechanisms~\cite{hu2018squeeze,cao2019gcnet,zhu2019empirical}.
In contrast, one-stage anchor-based detectors offer greater efficiency by eliminating the time-consuming proposal generation process.
SSD~\cite{liu2016ssd} scatters default boxes at different locations across multiple feature maps and directly uses a convolutional neural network to predict box offsets and category scores.
Subsequent one-stage detectors, such as RetinaNet~\cite{lin2017focal} and the YOLO series~\cite{redmon2017yolo9000,redmon2018yolov3,bochkovskiy2020yolov4} (except the original YOLO~\cite{redmon2016you}), follow this concept.
% As the first one-stage detector to employ the anchor notion, SSD~\cite{liu2016ssd} scatters default boxes at different locations across multiple feature maps and directly uses a convolutional neural network to predict box offsets and category scores.
% Thereafter, one-stage detectors progressively integrate anchor boxes into their designs for competitive performance as opposed to two-stage detectors, including RetinaNet~\cite{lin2017focal} and the YOLO series~\cite{redmon2017yolo9000,redmon2018yolov3,bochkovskiy2020yolov4} except the base YOLO~\cite{redmon2016you}.

However, the performance of anchor-based detectors with significant computational complexity is sensitive to the number, size and aspect ratio of anchors.
Consequently, recent research on anchor-free approaches has gained prominence and can be broadly categorized into two branches.
The first approach~\cite{law2018cornernet,zhou2019bottom}, based on keypoints, generates bounding boxes by identifying numerous pre-determined or self-learned keypoints.
The second approach~\cite{redmon2016you,kong2020foveabox,tian2020fcos}, center-based, utilizes the center point or region of objects to determine positive instances and predicts the four distances from the positives to the object border.

% Existing convnet-based object detectors can be classified into one-stage and two-stage ones according to whether they generate a series of region proposals before classification and regression. Two-stage detectors split detection problem into two sequential problems: first produce many regions of interest, then classify them and refine their locations. The R-CNN series~\cite{girshick2014rich,girshick2015fast,ren2015faster} are leading paradigm of two stage detectors. In contrast, one-stage detectors\cite{redmon2016you,liu2016ssd,lin2017focal} skip the proposal generation process and directly predict the class and bounding box at each location, thus with faster inference speed. However, most detectors are designed for human-collected benchmark with balanced and adequate training samples across all categories. Applying them to long-tailed datasets directly achieves inferior performance. Thus, we aim at improving detectors' performance in realistic scenarios.

\subsection{Long-Tailed Object Detection}
To handle the challenging long-tail problem in object detection, current approaches can be categorized into \textbf{data re-sampling}~\cite{shen2016relay,gupta2019lvis,chang2021image}, \textbf{specialized loss design}~\cite{cui2019class,tan2020equalization,tan2021equalization,wang2021adaptive,wang2021seesaw,feng2021exploring}, \textbf{exquisite architecture invention}~\cite{li2020overcoming,wu2020forest}, \textbf{decoupled training}~\cite{wang2020devil,li2020overcoming,zhang2021distribution,wang2021adaptive,feng2021exploring}, as well as \textbf{data augmentation}~\cite{ghiasi2021simple,zang2021fasa}.

\noindent\textbf{Data re-sampling.} Re-sampling is a popular solution to cope with imbalanced data, which involves under-sampling and over-sampling.
Nevertheless, under-sampling becomes infeasible when there is an extreme imbalance ratio between the head and tail categories.
On the contrary, over-sampling increases the occurrence frequency of samples from tail categories while retaining head samples.
Examples of over-sampling methods conducted at the image level include class-aware sampling~\cite{shen2016relay} and repeat factor sampling (RFS)~\cite{gupta2019lvis}.
At the object level~\cite{chang2021image}, a common memory mechanism is introduced to replay past features of tail categories, aiming to balance the sample distribution.
However, all these re-sampling techniques have limitations in enhancing sample diversity for tail categories.

\noindent\textbf{Specialized loss design.} Specialized loss design represents another prevalent approach for addressing the long-tail problem.
Cui {\em et al.}~\cite{cui2019class} propose the class balanced loss to adjust the loss proportion of assorted categories based on the effective number of samples.
Equalization loss (EQL)~\cite{tan2020equalization} creatively analyzes this issue from the perspective of loss gradients.
It truncates negative gradients from head classes to reduce suppression on the prediction for tail categories, at the cost of misclassifying instances from head categories as tail categories.
To improve the countermeasure, EQLv2~\cite{tan2021equalization} adaptively up-weights the positive gradients and down-weights the negative gradients of each category separately according to the 
accumulated gradient ratio of positives to negatives.
Additionally, ACSL~\cite{wang2021adaptive} only penalizes overconfident scores to prevent excessive
suppression on tail categories.
Seesaw loss~\cite{wang2021seesaw} dynamically calibrates the classification logits using complementary mitigation and compensation factors.
Likewise, LOCE~\cite{feng2021exploring} adaptively adjusts the classification logits based on the proposed mean classification score.
Notably, while these methods aim to simulate a more balanced training set by emphasizing the importance of tail classes, they often overlook the issue of limited sample diversity within tail categories.

\noindent\textbf{Exquisite architecture invention.} There have been efforts to modify the architecture of detectors in order to address the long-tail distribution.
Two typical representatives are BAGS\cite{li2020overcoming} and Forest R-CNN\cite{wu2020forest}, which aim to alleviate the dominance of head categories over tail categories by grouping all classes based on valuable prior knowledge.
Unfortunately, they prioritize addressing the unequal competition among foreground categories while overlooking the insufficient variation in samples from tail categories.

\noindent\textbf{Decoupled training.} Decoupled training separates the training process of the detector into two stages: representation learning and classifier learning.
Based on the findings~\cite{kang2019decoupling} that the long-tailed distribution is not an issue in learning high-quality representations, some works\cite{wang2020devil,li2020overcoming,zhang2021distribution,wang2021adaptive,feng2021exploring} apply their methods to the classifier learning stage in order to only adjust the classifier, while keeping other parameters frozen.
However, the frozen feature extractor in the classifier learning stage becomes the bottleneck for further improvements, especially when compared to the end-to-end approaches.

\noindent\textbf{Data augmentation.} As an effective approach to introducing extra sample variations, data augmentation is expected to supplement the undiversified tail categories in the context of long-tail distribution.
% However, data augmentation has been demonstrated to have the potential in promoting data diversity recently.
It turns out that appropriate data augmentation methods indeed bring further improvements to the long-tailed detection tasks.
A recently proposed augmentation strategy called simple Copy-Paste~\cite{ghiasi2021simple} has shown its superiority in improving feature representations for the decoupled training pipeline.
Besides, FASA~\cite{zang2021fasa} manages to address the data scarcity issue by augmenting samples in the feature space, especially for tail categories.
Nevertheless, it falls short of accurately capturing the feature distributions of all categories and involves a complex process to indirectly update the sampling probabilities of categories to be augmented.

\subsection{Feature hallucination}
Feature Hallucination is a widely adopted approach in zero-shot recognition tasks~\cite{zhu2020don,hayat2020synthesizing,zhao2020gtnet}.
These studies utilize a generative adversarial network~\cite{goodfellow2020generative} (GAN) conditioned on class semantic embeddings to synthesize features for unseen classes.
However, due to the inherent instability of GANs, training an auxiliary network to generate diverse features for tail categories is a non-trivial task.
In contrast, our proposed feature hallucination module involves no trainable parameters and prior knowledge such as class semantic embeddings.
Instead, it solely relies on the statistics of the feature extractor to consistently capture the precise feature distribution for each category during the detector's training process.
Simultaneously, it dynamically assigns higher selection probabilities to tail categories based on our proposed long-term indicators, and subsequently performs the feature synthesis process at each iteration.
Therefore, the feature hallucination module is highly efficient and can be seamlessly integrated into any detectors, providing real-time adjustment to the feature synthesis process.

% In light of the inadequacies of prior works, we offer a unified framework BACL that incorporates the re-balancing of training data distribution and the enhancement of data diversity for tail categories in the feature space.
% Motivated by this perspective, our proposed method not only re-balances the training distribution via a set of long-term and short-term indicators, but also manages to enrich the diversity of samples from extremely-few-shot tail categories by a novel feature hallucination module.

\section{Methodology}

\begin{figure*}
  \centering
  \includegraphics[width=\textwidth]{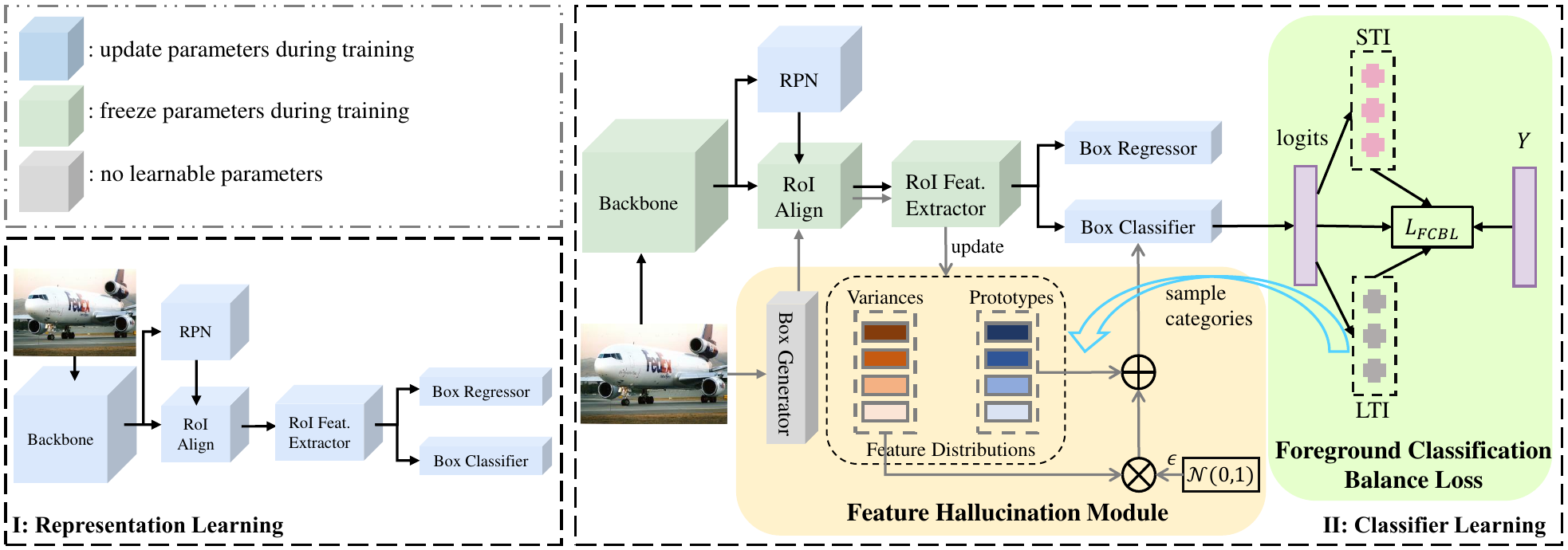}
  \caption{Overall framework of the proposed BACL. In the first stage (left panel), the entire object detector is trained on the long-tailed dataset for 12 epochs. In the second stage (right panel), we freeze the parameters of the feature extractor (green blocks), and fine-tune other parts (blue blocks) with our proposed foreground classification balance loss and feature hallucination module for another 12 epochs. Notably, LTI and STI stand for long-term indicators and short-term indicators, respectively.}
  \label{pic:3}
\end{figure*}

Conventional deep learning-based detectors often underperform, particularly on tail categories, in the presence of long-tailed detection datasets due to their inherent characteristics (in Section \ref{sec:3.1}).
To tackle this issue, we propose a decoupled training framework called BACL, as illustrated by Fig. \ref{pic:3}, which aims to obtain a generalized feature representation in the representation learning stage (in \ref{sec:3.2}) and calibrate the classification bias in the classifier learning stage (in \ref{sec:3.3}, \ref{sec:3.4} and \ref{sec:3.5}). 
In the second stage, to facilitate calibration, BACL freezes the parameters of the feature extractor within the detector and introduces a pair of complementary indicators for guidance (in \ref{sec:3.3}).
Building upon these indicators, BACL integrates a novel Foreground Classification Balance Loss (in \ref{sec:3.4}) to mitigate the unequal competition among various foreground categories.
Additionally, a dynamic Feature Hallucination Module (in \ref{sec:3.5}) is tailored to diversify the samples from tail categories.

% In this section, we start by giving a precise problem formulation of object detection and analyzing the limitations of Sigmoid Cross-Entropy loss when coping with long-tailed data (Section~\ref{sec:3.1}).
% Then, we introduce our proposed framework BACL (See Figure~\ref{pic:2}), which is guided by a novel long-short-term indicator (Section~\ref{sec:3.2}).
% Then, we elaborate on the components of BACL, including a novel Accuracy Balance Loss (Section~\ref{sec:3.3}) to alleviate the inconsistency of training and testing distributions and a dynamic Feature Hallucination Module to increase sample diversity (Section~\ref{sec:3.4}).

% In this section, we start by giving detailed formulation of long-tailed object detection (Section \ref{sec:3.1}). Then, we elaborate our proposed long-short-term indicator pair for reflecting the learning status of the classifier (Section \ref{sec:3.2}). Based on this pair, we propose \textbf{B}a\textbf{LAN}ced \textbf{C}lassification (\textbf{BLANC}), a unified framework for tackling classification bias, to tackle consisting of two components: 1) a novel indicator-pair-guided accuracy balance loss (\textbf{ABL}, Section \ref{sec:3.3}) that provides positive margins for samples from the weak class between any two classes, especially confusing classes ({\em e.g.}, bicycle vs. tricycle), 2) a dynamic \textbf{F}eature \textbf{H}allucination \textbf{M}odule (\textbf{FHM}, Section \ref{sec:3.4}) that generates extra training examples in the region of interest (RoI) feature space based on high quality feature prior per class.

\subsection{Preliminary}\label{sec:3.1}

Following pioneering works~\cite{li2020overcoming, wang2021adaptive}, we adopt the popular two-stage object detector Faster R-CNN to implement our proposed methods.
In the Faster R-CNN pipeline, the backbone network randomly takes an image as input and generates a corresponding feature map.
RPN performs convolution on the feature map and produces a fixed number of region proposals.
Next, the RoIAlign layer first pools these region proposals into fixed size according to the feature map and then the RoI feature extractor, consisting of two fully connected (FC) layers, encodes them into $d$-dimensional RoI features $\bm{h}$.
Finally, these RoI features are fed to two separate FC layers dedicated to classification and regression.
In the classification branch, the network is tasked with solving a $(C+1)$-category prediction problem ($C$ foreground categories and the background) by applying $\bm{z}=FC_{cls}(\bm{h})$, where $\bm{z}=[z_1,\cdots, z_{C+1}]$ represents the predicted logits.

Traditionally, the classification branch is supervised by the softmax cross-entropy loss during training.
For a proposal labeled as $i$, the gradient passed by the loss for each category $j$ is formulated as follows:
\begin{equation}
    \frac{\partial L_{cls}}{\partial z_j}=\left\{
        \begin{aligned}
            & p_j - 1, & j=i\\
            & p_j, & j\neq i
        \end{aligned}
    \right.
\end{equation}
In the long-tailed setting, when categories $i$ and $j$ belong to the head and tail categories, respectively, category $j$ receives overwhelmingly large suppression gradients from category $i$, as proved by~\cite{tan2020equalization,wang2021seesaw}.
Therefore, the classifier often assigns low probabilities to the tail category $j$ even when presented with positive proposals, leading to low classification accuracy.

In order to improve classification accuracy, previous studies~\cite{tan2020equalization,wang2021adaptive,tan2021equalization} have adopted a sigmoid-based classifier that has demonstrated superiority in large vocabulary datasets.
EQLv2 introduces an additional objectness branch to reduce false positives, considering all proposals belonging to the background as positive samples.
Mathematically, the classification loss for a training sample with logits $\bm{z}=[z_1,\cdots, z_{C+1}]$ and one-hot labels $\bm{y}=[y_1,\cdots, y_{C+1}]$ is formulated as follows:
\begin{equation}\label{eqn:2}
    L_{cls} = -\sum_{i=1}^{C+1}\log(\hat{p_i}),
\end{equation}
where,
\begin{equation}\label{eqn:3}
    \hat{p_i} = \left\{
        \begin{aligned}
            &p_i, & y_i=1 \\
            &1-p_i, & y_i=0
        \end{aligned}
    \right.
\end{equation}
\begin{equation}\label{eqn:4}
    p_i = \frac{1}{1+\exp(-z_i)}.
\end{equation}
In the inference phase, the estimated probability vector $\tilde{\bm{p}}=[\tilde{p}_1, \tilde{p}_2, \cdots, \tilde{p}_{C+1}]$ of the $(C+1)$-channel sigmoid-based classifier can be expressed by the following equation:
\begin{equation}\label{eqn:5}
    \tilde{p}_i = \left\{
        \begin{aligned}
            & (1-p_{C+1})\cdot p_i, & 1 \le i\le C\\
            & p_{C+1}, &i=C+1
        \end{aligned}
    \right.
\end{equation}

\subsection{Representation Learning Stage}\label{sec:3.2}

BACL focuses on obtaining a generalized feature representation during the representation learning stage, which serves as a robust foundation for the subsequent classifier learning stage.
To this end, we improve the training strategy of Faster R-CNN in the representation learning stage through comprehensive experimentation and observations.
The quantitative results for each improvement are presented in Tab. \ref{tab:representation_learning}.
For evaluation, we adopt 101-point interpolated average precision for box predictions $AP^{b}$ over 10 IoU thresholds ranging from 0.5 to 0.95 and all classes.

\subsubsection{Sigmoid-based Classifier with an Objectness Branch}

As demonstrated by~\cite{zhou2020bbn}, the use of re-balancing methods can unexpectedly degrade the feature representation.
Therefore, we employ a sigmoid-based classifier with an objectness branch, as stated in \ref{sec:3.1} and formulated in Eq.~\eqref{eqn:2}-\eqref{eqn:5}, without applying any re-balancing techniques to adjust the distribution.
% Motivated by~\cite{tan2021equalization}, we adopt the sigmoid-based classifier to facilitate the subsequent design of FCBL and add an objectness branch to reduce false positives that .
% Hence, for a training sample with logits $\bm{z}=[z_1,\cdots, z_{C+1}]$ and one-hot labels $\bm{y}=[y_1,\cdots, y_{C+1}]$, the classification loss is formulated as:
% \begin{equation}\label{eqn:2}
%     L_{cls} = -\sum_{i=1}^{C+1}\log(\hat{p_i}),
% \end{equation}
% where,
% \begin{equation}\label{eqn:3}
%     \hat{p_i} = \left\{
%         \begin{aligned}
%             &p_i, & y_i=1 \\
%             &1-p_i, & y_i=0
%         \end{aligned}
%     \right.
% \end{equation}
% \begin{equation}\label{eqn:4}
%     p_i = \frac{1}{1+\exp(-z_i)}.
% \end{equation}
% In the inference phase, the estimated probability vector $\tilde{\bm{p}}=[\tilde{p}_1, \tilde{p}_2, \cdots, \tilde{p}_{C+1}]$ of the $(C+1)$-channel sigmoid-based classifier satisfies the following equation:
% \begin{equation}\label{eqn:5}
%     \tilde{p}_i = \left\{
%         \begin{aligned}
%             & (1-p_{C+1})\cdot p_i, & 1 \le i\le C\\
%             & p_{C+1}, &i=C+1
%         \end{aligned}
%     \right.
% \end{equation}
% \begin{align}
%     \tilde{p}_i = (1-\tilde{p}_{C+1})&\cdot \tilde{p}_i,\; i\in{1,2,\cdots, C}\\
%     \tilde{p}_{C+1} &= p_{C+1}
% \end{align}

\subsubsection{Leverage the Simple Copy-Paste Augmentation~\cite{ghiasi2021simple}}
It is widely acknowledged that data augmentation methods can enhance the feature representation by introducing additional variations in the data. With this in mind, we replace the conventional multi-scale training strategy with the simple Copy-Paste augmentation strategy, which has the capability to create more challenging training samples, thereby leading to better feature representations for the decoupled training.
% The simple Copy-Paste augmentation strategy randomly selects a subset of objects from one image and pastes them onto another image after applying standard scale jittering and random horizontal flipping to both images.
% Such an easy-to-plug-in data augmentation strategy has demonstrated its effectiveness on multiple benchmarks.
% Particularly, it results in better feature representations for the two-stage training procedure typically used in the LVIS benchmark, which is in par with our goal.
% Therefore, we leverage the simple Copy-Paste augmentation strategy instead of the commonly adopted multi-scale training strategy during the representation learning stage, providing a more generalized feature extractor for the classifier learning stage.

\subsubsection{Other Feasible Attempts}

Through extensive experiments, we observe that reducing the weight decay coefficient slightly improves representation learning. This can be attributed to the fact that decreasing the restriction on weight magnitude expands the parameter search space for feature expression, ultimately leading to enhanced representations.
In addition, by doubling the number of proposals retained after the Non-Maximum Suppression operation from 1000 to 2000, we increase the number of foreground proposals that are forwarded to the subsequent RoI feature extractor.
As a consequence, this increase in quantity aids in the convergence of the RoI feature extractor and yields slightly improved feature representations.
% In addition, doubling the number of proposals kept after Non-Maximum Suppression operation contributes to performance gains as well.

\begin{table*}[t]
    \centering
    \caption{Gradual performance improvements on LVIS v0.5 \texttt{val} set during the representation learning stage. $AP^b$ denotes the 101-point interpolated average precision  for box predictions over 10 IoU thresholds ranging from 0.5 to 0.95 and all classes, while $AP_r$, $AP_c$, and $AP_f$ represent the detection average precision for rare, common, and frequent categories, respectively.}
    \begin{tabular}{c|ccccc|cccc}
    \toprule
    Model & Sigmoid & Objectness Branch & Double Proposals & Small Weight Decay & Simple Copy-Paste & $AP^b$ & $AP_r$ & $AP_c$   & $AP_f$   \\
    \midrule
    \multirow{6}*{Faster R-CNN} & \ding{55} & \ding{55} & \ding{55} & \ding{55} & \ding{55} & 18.0 & 1.6 & 15.0 & 28.3 \\
    ~ & \ding{51} & \ding{55} & \ding{55} & \ding{55} & \ding{55} & 18.7 & 2.4 & 16.8 & 27.7 \\
    ~ & \ding{51} & \ding{51} & \ding{55} & \ding{55} & \ding{55} & 20.5 & 3.6 & 18.8 & 29.4 \\
    ~ & \ding{51} & \ding{51} & \ding{51} & \ding{55} & \ding{55} & 20.6 & 4.2 & 18.8 & 29.5 \\
    ~ & \ding{51} & \ding{51} & \ding{51} & \ding{51} & \ding{55} & 21.2 & 4.9 & 20.0 & 29.2 \\
    ~ & \ding{51} & \ding{51} & \ding{51} & \ding{51} & \ding{51} & \textbf{21.7} & \textbf{4.9} & \textbf{20.4} & \textbf{30.1} \\
    \bottomrule
    \end{tabular}
    \label{tab:representation_learning}
\end{table*}

\begin{table*}[t]
    \centering
    \caption{Different types of long-term indicators. The constant column indicates whether this kind of indicators will alter during calibration. In the criterion column, we assume that category $i$ is stronger than $j$. In the equation column, $y_{n,i}$ is the one-hot label of category $i$ for the $n$-th sample and $\mathbb{I}[\cdot]$ is an indicator function that outputs 1 if the input condition holds, otherwise 0.}
    \begin{tabular}{c|ccccc}
    \toprule
    Types & Name & Equation\centering & Constant & Dimension & Criterion \\
    \midrule
    \multirow{2}{*}{Static Stat.} & image frequency & $f_i$ & \ding{51} & 1 & $f_i > f_j$ \\
    ~ & instance frequency & $F_i$ & \ding{51} & 1 & $F_i > F_j$ \\
    \midrule
    \multirow{6}{*}{First-Order Dynamic Stat.} & cumulative instance number & $N_i = \sum\limits_n \mathbb{I}[y_{n,i}=1]$ & \ding{55} & 1 & $N_i>N_j$ \\
    ~ & mean classification score & $s_i^t = \gamma s_i^{t-1} + (1 - \gamma) p_i^t$ & \ding{55} & 1 & $s_i^t>s_j^t$ \\
    \rule{0pt}{20pt}
    ~ & true positive rate & $T\!P\!R_i = \frac{\sum\limits_n \mathbb{I}[\mathop{\mathrm{argmax}}\limits_k p_{n,k}=i]\cdot\mathbb{I}[y_{n,i}=1]}{\sum\limits_n \mathbb{I}[y_{n,i}=1]}$ & \ding{55} & 1 & $T\!P\!R_i>T\!P\!R_j$ \\
    \midrule
    \multirow{3}{*}{Second-Order Dynamic Stat.} & \multirow{3}{*}{confusion matrix} & 
    $M_{i,j} = \frac{\sum\limits_n \overline{p}_{n,j}\cdot\mathbb{I}[y_{n,i}=1]}{\sum\limits_n \mathbb{I}[y_{n,i}=1]},$
    & \multirow{3}{*}{\ding{55}} & \multirow{3}{*}{2} & \multirow{3}{*}{$M_{j,i}>M_{i,j}$} \\
    ~ & ~ & $\overline{p}_{n,j}=\frac{\exp(z_j)}{\sum_{k=1}^{C}\exp(z_k)}$ & ~ & ~ & ~\\
    \bottomrule
    \end{tabular}
    \label{tab:long-term indicator}
\end{table*}

\subsection{The long-short-term indicators pair}\label{sec:3.3}

To facilitate calibrating the classification bias in the classifier learning stage, we introduce a pair of complementary long-term and short-term indicators to reflect the learning status of the classifier from two perspectives: the inclination towards classifying foreground categories and the correctness of classification for each category.
% To get rid of adverse effects of label distribution shift, we propose a long-short-term indicator pair to monitor equilibrium and correctness of the classifier respectively. 

On the one hand, the long-term indicators should accurately capture the dominance among foreground categories and the classifier's inclination towards various foreground categories during training, so that they can serve as a guide to achieve classification equilibrium.
Three types of statistics fulfill this requirement: static statistics, first-order dynamic statistics, and second-order dynamic statistics.
Each type is enumerated in Tab. \ref{tab:long-term indicator}, along with the corresponding criterion for determining the dominant category assuming category $i$ is dominant over category $j$.

Firstly, static statistics provide prior information about the dataset and remain constant throughout training.
However, they cannot dynamically track classification equilibrium and only offer a rough estimate.
The most commonly used static statistics include the one-dimensional image frequency $f_i$ and instance frequency $F_i$ for each category $i$.

Secondly, first-order dynamic statistics provide a better insight into the state of classification equilibrium as they are continuously updated during training.
These statistics include the cumulative instance number $N_i$, mean classification score~\cite{feng2021exploring} $s_i^t$ at iteration $t$, and online true positive rate $T\!P\!R_i$ for each category $i$
(refer to Tab. \ref{tab:long-term indicator} for concrete equations).

Thirdly, second-order dynamic statistics, represented by the confusion matrix~\cite{he2022relieving} $M\in\mathbb{R}^{C\times C}$, capture fine-grained inter-class relationships.
They provide a theoretically superior approach compared to statistics that only consider classification information within each category.

On the other hand, while the aforementioned long-term indicators are effective in indicating classification equilibrium, they lack the ability to assess correctness.
To address this limitation, we introduce short-term indicators that can evaluate the correctness of classification results.
These indicators coach the classifier to focus on challenging categories, thereby further enhancing discrimination power.
Specifically, the first $C$ terms of the predicted probability vector $\tilde{\bm{p}}$ in Eq. \eqref{eqn:5} seamlessly fulfill our requirements.
Hence, in this work, we consider the vector $[\tilde{p}_1, \tilde{p}_2, \cdots, \tilde{p}_{C}]$ as the short-term indicators.

In summary, long-term and short-term indicators are complementary, providing orthogonal viewpoints for monitoring the learning status of the classifier.
They address the shortcomings of previous works that focused solely on either aspect, forming the foundation of our proposed FCBL and FHM.

\subsection{Formulation of Foreground Classification Balance Loss}\label{sec:3.4}

On the basis of the complementary indicators pair, we devise a Foreground Classification Balance Loss (FCBL) to tackle the prevailing unequal competition among diverse foreground categories in the long-tail scenarios. It is noteworthy that this loss function is exclusively applied to foreground proposals, whereas the loss of background proposals remains to be calculated by Eq. \eqref{eqn:2}.
This design allows the classifier to maintain discrimination power between foreground and background.
Formally, FCBL can be expressed as follows:
\begin{equation}\label{eqn:11}
    L_{FCBL} = -\log(p_i)-\log(1-p_{C+1})-\sum\limits_{j=1,j\neq i}^{C}w_j\log(1-p_{j}^{\prime}),
\end{equation}
where $i$ is the ground-truth label for each foreground proposal, and $p_i$, $p_{C+1}$ are in consistency with Eq. \eqref{eqn:4}. However, $p_j^{\prime}$ is given by Eq. \eqref{eqn:13}; $w_j$ is defined as specified in Eq. \eqref{eqn:14}.

Firstly, FCBL introduces an adaptive class-aware margin between any pair of foreground categories, in order to ameliorate the domination of one category over another.
The margin is logarithmically proportional to the ratio of the corresponding long-term indicators.
For example, considering a sample from foreground category $i$, the margin $\delta_{ij}$ between $i$ and another foreground category $j$ is defined by Eq. \eqref{eqn:12}:
\begin{equation}\label{eqn:12}
    \delta_{ij} = \alpha\cdot\log(\frac{l_j}{l_i})
\end{equation}
where $\alpha$ controls the range of the margin, and $l_i/l_j$ represents a unified expression for long-term indicators, which can take the form of $f_i$, $F_i$, $N_i$, $s_i^t$, $T\!P\!R_i$, or $M_{j,i}$ for $l_i$.

The probability for each non-ground-truth foreground category $j$ is then reformulated according to Eq. \eqref{eqn:13}:
\begin{equation}\label{eqn:13}
    p_{j}^{\prime} = \frac{1}{1+\exp[-(z_j+\delta_{ij})]}
\end{equation}
In this form, the adaptive class-aware margin $\delta_{ij}$ possesses the following characteristics:
(1) If the ground-truth category $i$ is stronger than category $j$ (as determined by the criterion in Tab. \ref{tab:long-term indicator}), the margin $\delta_{ij}$ will be negative.
This reduction in suppression allows the classifier to assign a higher probability to category $j$.
(2) Conversely, if the condition is reversed, $\delta_{ij}$ will be positive. 
This positive margin encourages the classifier to decrease the confidence of the strong category $j$ through a larger suppression gradient.
Therefore, this margin design enables FCBL to perceive category discrepancies and dynamically adjust the magnitude of suppression gradients.

\begin{figure}
    \centering
    \includegraphics[width=\columnwidth]{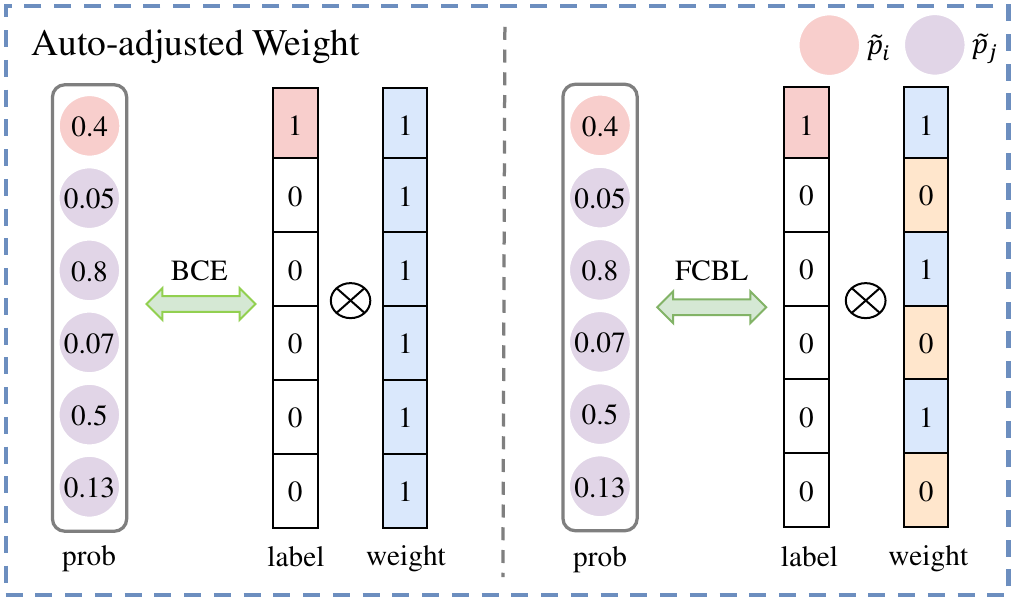}
    \caption{Visualization of the auto-adjusted weight in FCBL. In default, the tunable threshold $\tilde{p}_{t}$ is set to 0.7.}
    \label{pic:4}
\end{figure}

Secondly, natural data and long-tailed datasets usually possess a large vocabulary set, which raises the difficulty of training the classifier without bias.
Motivated by ACSL~\cite{wang2021adaptive}, FCBL incorporates an auto-adjusted weight term $w_j$, defined by Eq. \eqref{eqn:14}, into the binary cross-entropy loss of each non-ground-truth foreground category $j$ (where $1\le j \le C$ and $j\neq i$), where $\tilde{p}_{t}$ is a pre-defined threshold and $\tilde{p}_{i}$, $\tilde{p}_{j}$ represent the corresponding short-term indicators.
\begin{equation}\label{eqn:14}
    w_{j}=\left\{
        \begin{aligned}
            & 1, \quad & \tilde{p}_{j}\geq \tilde{p}_{i} \\
            & 1, \quad & \tilde{p}_{j}\geq \tilde{p}_{t} \\
            & 0, \quad & \text{otherwise}
        \end{aligned}
    \right.
\end{equation}
The introduction of the auto-adjusted weight term aims to prioritize confounding categories while disregarding well-classified ones, as depicted by Fig. \ref{pic:4}.
Specifically, learning a suitable mapping from output logits to concrete category predictions warrants the losses of misclassified categories ($\tilde{p}_{j}\geq \tilde{p}_{i}$).
% In detail, losses of those misclassified categories ($\tilde{p}_{j}\geq \tilde{p}_{i}$) are necessary because they force the classifier to learn a suitable mapping from output logits to concrete category predictions.
Besides, although some non-ground-truth categories may not have higher probabilities than the actual category, the classifier still allocates excessively high probabilities over a pre-defined threshold to them ($\tilde{p}_{j}\geq \tilde{p}_{t}$).
This observation suggests that the classifier lacks discrimination power between the ground-truth category and these similar, non-ground-truth categories.
By aggregating losses from these two sources during backpropagation, the classifier's generalization ability can be improved.
Regrettably, ACSL, which shares a similar motivation, only considers the latter scenario.

Finally, in the inference phase, the estimated probability vector for the classifier learning stage remains as $\tilde{\bm{p}}=[\tilde{p}_1, \tilde{p}_2, \cdots, \tilde{p}_{C+1}]$, which is defined by Eq. \eqref{eqn:5}.

\begin{algorithm}[t]
    \caption{The pipeline of the classifier learning stage}
    \label{alg:algorithm}
    \textbf{Input}: Batch annotated images $\mathcal{I}$ with $N_I$ objects belonging to $C_I$ classes
    % \textbf{Parameter}: The RPN, a Box Generator, the feature extractor $\mathcal{F}$, the box classifier $\mathcal{C}$, the box regressor $\mathcal{R}$

    \begin{algorithmic}[1] %[1] enables line numbers
    \STATE \textbf{Initialize:} frozen feature extractor $\mathcal{H}$ ({\em i.e.}, backbone, RoIAlign layer and RoI feature extractor) from the representation learning stage
    \FOR{1,$\cdots$,$N_I$}
    \STATE $\hat{b}=[x_1+\frac{\eta_1 w}{6},y_1+\frac{\eta_2 h}{6}, x_2+\frac{\eta_3 w}{6},y_2+\frac{\eta_4 h}{6}]$  $\vartriangleright$ boxes transformation with $b_{GT}=[x_1,y_1,x_2,y_2]$ and label $i$
    \STATE $h=\mathcal{H}(\hat{b})$
    \ENDFOR
    \FOR{$i$ in 1,$\cdots$,$C_I$}
    \STATE Calculate $u_{i}$ and $v_{i}$ of all features $h$ with label $i$
    \STATE $\mu_{i}\leftarrow \beta\mu_{i}+(1-\beta)u_i,\quad \sigma_{i}\leftarrow \beta\sigma_{i}+(1-\beta)v_i$ $\vartriangleright$ $\mu_{i}$, $\sigma_{i}$ are the prototype and variance of class $i$
    \ENDFOR $\quad\vartriangleright$ feature distributions alteration
    \STATE Select $c$ classes with probabilities $sp_i$ in Eq. \eqref{eqn:17}
    \FOR{$i$ in 1,$\cdots$,$c$}
    \STATE Repeat $f_i=\mu_i+\epsilon\odot\sigma_i, \quad \epsilon\in \mathcal{N}(0,1)$ for $m$ times
    \ENDFOR $\quad\vartriangleright$ hallucinated features synthesization
    \STATE Update long-short-term indicators pair in Tab. \ref{tab:long-term indicator} and Eq. \eqref{eqn:5} with hallucinated features and RoI features from RPN
    \vspace{-4mm}
    \STATE Calculate classification loss with Eq. \eqref{eqn:11}
    \end{algorithmic}
    \textbf{Output}: The classification loss $L_{FCBL}$
\end{algorithm}

\subsection{Feature Hallucination Module}\label{sec:3.5}

While FCBL addresses the issue of unequal competition among foreground categories, it fails to resolve the problem of under-representation in tail categories.
For instance, the category "bait" has only one training sample.
The extreme scarcity of training samples hinders the classifier from possessing sufficient discrimination power for this category.
Most previous studies have employed various re-sampling methods to increase the number of training samples for tail categories.
However, sample diversity highlights not only the quantity but also the variations among samples.
Unfortunately, prior approaches have neglected to incorporate additional data variations, thereby hindering further improvement in tail categories.
% The majority of earlier studies have used different re-sampling methods to boost the number of training samples from tail categories, but fail to introduce extra data variances.

To overcome this problem, we offer a straightforward but effective feature hallucination module (FHM), which enhances the representation in the feature space, especially for tail categories, by synthesizing hallucinated features to boost data diversity.
As illustrated in Fig. \ref{pic:3}(b), FHM initially captures the feature distribution of each category in real-time and then generates training features for selected categories based on the guidance provided by the long-term indicators.

% For every category $j$, FHM maintains its feature prior online, including the prototype $\mu_{j}$ and the dispersion $\sigma_{j}$, updating them with high-quality roi features via the exponential moving average function~\eqref{equ:2}. Based on the feature prior, FHM generates extra hallucinated roi features of selected categories by the reparametrization trick to compensate for low diversity.

Concretely, FHM generates region proposals that exhibit substantial overlaps with the ground-truth bounding boxes using a non-learnable bounding box generator, as inspired by~\cite{feng2021exploring}.
In contrast to RPN, the bounding box generator uses coordinate manipulation to stochastically transform the ground-truth bounding boxes in an image into positive proposals.
Formally, for a batch of training images $I$, the designed box generator takes the box coordinates $b_{GT}=[x_1,y_1,x_2,y_2]$ of each instance in $I$ as input, where $(x_1,y_1)$, $(x_2,y_2)$ represent the coordinates of the top-left and bottom-right corners, respectively.
It then generates 16 dense proposals for each instance in $I$, with slight offsets in their locations, as follows:
\begin{equation}
    \hat{b}=[x_1+\frac{\eta_1 w}{6},y_1+\frac{\eta_2 h}{6}, x_2+\frac{\eta_3 w}{6},y_2+\frac{\eta_4 h}{6}],
\end{equation}
where $\eta_{i}\in[-1,1]$ is randomly selected and $w,h$ are box width and height, {\em i.e.}, $w=x_2-x_1,h=y_2-y_1$.
Clearly, each proposal is slightly different from the remaining 15 ones in terms of foreground and background, which is conducive to increasing sample variances.

Afterward, the subsequent RoIAlign layer and RoI feature extractor encode them as RoI features, not for classification and regression, but for collecting online feature distributions, which include prototypes and variances.
In detail, FHM computes the mean $u_i$ and variance $v_i$ of features for each category $i$ that emerges in $I$ and then alters the corresponding prototype $\mu_i$ and variance $\sigma_i$ using the exponential moving average function defined in Eq. \eqref{eqn:16}.
\begin{equation}
    \begin{aligned}\label{eqn:16}
        \mu_{i}\leftarrow \beta\mu_{i}+(1-\beta)u_i,\\
        \sigma_{i}\leftarrow \beta\sigma_{i}+(1-\beta)v_i.
    \end{aligned}
\end{equation}
% To be more precise, we first derive proposals with high IoU scores from a Box Generator~\cite{feng2021exploring} to acquire high-quality RoI features. The designed box generator takes an object of category $j$ with the ground-truth box $b_{GT}=[x_1,y_1,x_2,y_2]$ as input, then outputs 16 dense bounding boxes with slight offsets at iteration $t$:
% \begin{equation}
%     \hat{b}=[x_1+\frac{\eta_1 w}{6},y_1+\frac{\eta_2 h}{6}, x_2+\frac{\eta_3 w}{6},y_2+\frac{\eta_4 h}{6}],
% \end{equation}
% where $\eta_{i}\in[-1,1],w=x_2-x_1,h=y_2-y_1$. Feeding these boxes into subsequent RoI align module and RoI feature extractor, we obtain high-quality features with little noise.

Lastly, FHM guarantees the prominence of tail categories by allocating a sampling probability $sp_i$ to each category $i$, which is inversely proportional to the long-term indicator $l_i$:
\begin{equation}\label{eqn:17}
    sp_i=\frac{1-l_i}{\sum\limits_{k=1}^{C}(1-l_k)}, \; i \in1,\cdots C,
\end{equation}
where $l_i$ can take the form of $f_i$, $F_i$, $N_i$, $s_i^t$, $T\!P\!R_i$, or $M_{i,i}$.
Utilizing the aforementioned sampling probabilities, FHM randomly selects $c$ categories and generates $m$ hallucinated features for each category $i$ by the constantly renewed feature distribution through the reparametrization trick~\cite{kingma2013auto}:
\begin{equation}
    f_i=\mu_i+\epsilon\odot\sigma_i, \; \epsilon\in \mathcal{N}(0,1).
\end{equation}

Therefore, guided by the long-term indicators, FHM dynamically intensifies the data diversity by introducing novel hallucinated features, particularly for tail categories, thereby mitigating the issue of under-representation.

In conclusion, the pipeline of the classifier learning stage in the proposed BACL framework is outlined in Algorithm \ref{alg:algorithm}.

% Last but not least, we design a sampler with probability $sp$ reverse to the long-term indicator $CCA$ to decide which categories to :
% \begin{equation}
%     sp_j=\frac{1-CCA_j}{\sum\limits_{l=1}^{C}1-CCA_l}, \quad j \in\{1,\cdots C\}.
% \end{equation}
% Therefore, weaker classes have higher probabilities to be chosen for feature hallucination, so as to alleviate data scarcity of tail categories more frequently. Every iteration, we select $k$ categories according to the sampler, then use the reparametrization trick to generate $m$ hallucinated features per category to join the subsequent training.

\section{Experiments}

\subsection{Dataset and Evaluation Metric}

We conduct extensive experiments on the long-tailed Large Vocabulary Instance Segmentation (LVIS) dataset to confirm the efficacy of our proposed BACL framework.
LVIS serves as a challenging benchmark for long-tailed object recognition and is available in two versions: v0.5 and v1.0.
The \texttt{train} set and \texttt{val} set of LVIS v0.5 comprise 1230 and 830 categories, respectively, while the \texttt{test} set remains unidentified.
The three sets (57k \texttt{train}, 5k \texttt{val}, and 20k \texttt{test}) contain a total of 82k annotated images.
However, LVIS v1.0 includes 1203 and 1035 categories in the \texttt{train} set and \texttt{val} set, which provides a split of \texttt{train}, \texttt{val}, and \texttt{test} with 100k, 19.8k, and 19.8k images, respectively.
In both versions, LVIS categorizes the classes in the \texttt{train} split into rare ($\leq 10$ images), common (11-100 images), and frequent ($\geq 100$ images) groups based on the number of images.

We adopt 101-point interpolated average precision for box predictions $AP^{b}$ over 10 IoU thresholds ranging from 0.5 to 0.95 and all classes.
In addition, we also report $AP_{r}$, $AP_{c}$, $AP_{f}$ to indicate the detection performance of rare, common, and frequent categories, respectively.
Remarkably, all the aforementioned metrics disregard the results for categories that are absent in the \texttt{val} set due to the sparse annotation of LVIS.

\subsection{Implementation Details}

We implement our proposed framework with the open-source toolbox MMDetection~\cite{chen2019mmdetection} and conduct experiments primarily on the standard Faster R-CNN~\cite{ren2015faster}.
In terms of the model architecture, we opt for the popular ResNet~\cite{he2016deep} with FPN~\cite{lin2017feature} as the backbone, which is initialized by the ImageNet pre-trained model.
As mentioned earlier, we adopt the decoupled training pipeline.
Specifically, the model is initially trained for 12 epochs in the representation learning stage. Subsequently, the parameters of the backbone, RoIAlign layer, and RoI feature extractor are frozen, and the model is fine-tuned for another 12 epochs in the classifier learning stage.
The models are trained with a total batch size of 16 on 8 GPUs.
In each stage, the learning rate starts at 0.02 and decreases by 0.1 after 8 and 11 epochs, respectively.
Throughout the training process, we increase the number of proposals retained after non-maximum suppression from 1000 to 2000.

In the representation learning stage, the optimizer is SGD with momentum 0.9 and weight decay 0.00005 instead of 0.0001.
This choice is based on the superiority of a smaller weight decay in feature extraction, as mentioned in Section \ref{sec:3.2}.
The classification loss used during training is defined in Eq. \eqref{eqn:2}.
Instead of using standard scale jittering, we apply simple Copy-Paste and random horizontal flipping on images with a size of 1280$\times$1280.

% Following the convention, images are resized such that the shorter edge is within $\{640,672,704,736,768,800\}$ and the longer edge is less than 1333 pixels with only random horizontal flipping. We adopt a decoupled training pipeline to train the detector for 24 epochs with a total batch size 8 on 4 GPUs. Namely, we first train the detector for 12 epochs, then finetune it with proposed components for another 12 epochs. During each process, the learning rate starts at 0.01 and decreases by 0.1 after 8 and 11 epochs, respectively. In reference, no data augmentation is adopted and we keep the top 300 detections as the final results with the minimum score threshold of $10^{-3}$.

While in the classifier learning stage, the weight decay is set back to 0.0001.
Following common practice, we apply random horizontal flipping and scale jittering such that the shorter edge of the image is within $\{640,672,704,736,768,800\}$ pixels while the longer edge is less than 1333 pixels.
As for the hyperparameters, we set $\gamma$ as 0.9 to update the mean classification score following \cite{feng2021exploring}.
In the FCBL formulation, we set $\alpha$ to 0.85 to control the magnitude of the pairwise class-aware margin, and the probability threshold $\tilde{p}_t$ is set to 0.7.
Additionally, we configure $\beta$ as 0.9 to capture feature distributions in FHM. 
The number of sampled categories $c$ and synthetic hallucinated features $m$ are 8, 12, separately.

At testing time, we adopt single-scale testing with the image size of $1333\times800$ pixels without any test time augmentation for a fair comparison.
Following~\cite{gupta2019lvis}, we decrease the score threshold from 0.05 to 0.0001 and raise the maximum number of predictions per image from 100 to 300 after applying non-maximum suppression  with an IoU threshold of 0.5.

\begin{figure}[t]
    \centering
    \includegraphics[width=\columnwidth]{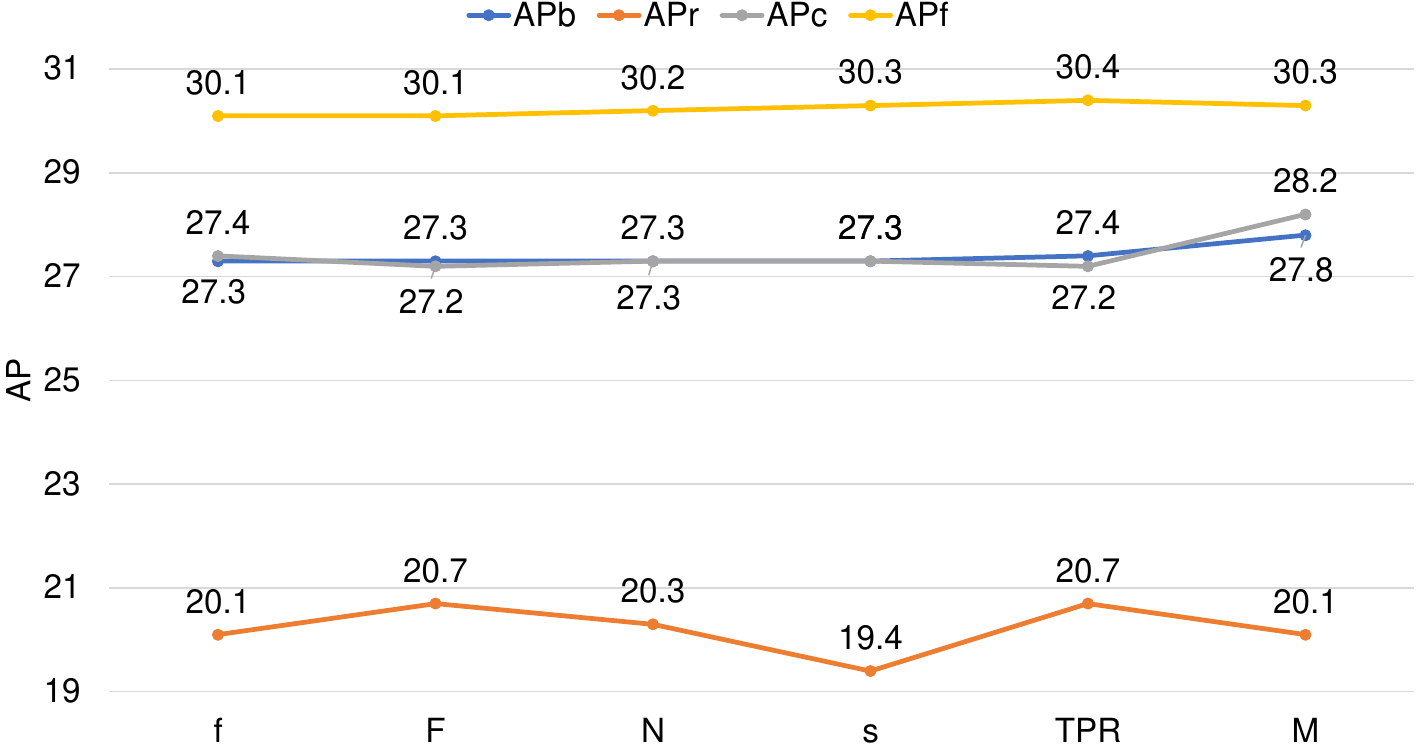}
    \caption{Performance comparison of different long-term indicators as listed in Tab. \ref{tab:long-term indicator} on the challenging LVIS v0.5.}
    \label{pic:5}
\end{figure}

\begin{table}[t]
    \renewcommand\arraystretch{0.9}
    \centering
    \caption{Results for all arbitrary combinations of three components. MG, WT, and FHM represent the class-aware margin, weight term, and feature hallucination module, respectively.}
    \begin{tabular}{c|ccc|c|ccc}
    \toprule
    Method & MG & WT & FHM & $AP^b$ & $AP_r$ & $AP_c$   & $AP_f$   \\
    \midrule
    Baseline &  &  &  & 22.0 & 4.0 & 21.2 & 30.1 \\
    - & \checkmark &  &  & 24.3 & 8.7 & 24.7 & 30.1 \\
    - &  & \checkmark &  & 24.0 & 8.5 & 23.8 & 30.3 \\
    FHM &  &  & \checkmark & 26.5 & 17.9 & 26.7 & 29.8 \\
    FCBL & \checkmark & \checkmark &  & 24.2 & 8.5 & 24.5 & 30.3 \\
    - & \checkmark &  & \checkmark & 26.7 & 18.3 & 27.1 & 29.6 \\
    - &  & \checkmark & \checkmark & 27.2 & 20.7 & 26.9 & 30.2 \\
    BACL & \checkmark & \checkmark & \checkmark & \textbf{27.8} & \textbf{20.1} & \textbf{28.2} & \textbf{30.3} \\
    \bottomrule
    \end{tabular}
    \label{tab:ca}
\end{table}
  
\begin{table}[t]
        \renewcommand\arraystretch{0.9}
        \centering
        \caption{Results for different values of $\alpha$ in FCBL.}
        \begin{tabular}{c|c|ccc}
        \toprule
        $\alpha$ & $AP^b$  & $AP_r$  & $AP_c$  & $AP_f$  \\
        \midrule
        0 & 27.2 & 20.7 & 26.9 & 30.2 \\
        % 0.6 & 27.5 & 20.7 & 27.5 & 30.1 \\
        0.7 & 27.4 & 21.0 & 27.2 & 30.2 \\
        0.8 & 27.5 & 21.0 & 27.6 & 30.1 \\
        \textbf{0.85} & \textbf{27.8} & \textbf{20.1} & \textbf{28.2} & \textbf{30.3} \\
        0.9 & 27.6 & 21.3 & 27.5 & 30.2 \\
        1.0 & 27.5 & 21.3 & 27.4 & 30.2 \\
        % 1.1 & 27.4 & 20.9 & 27.3 & 30.2 \\
        \bottomrule
        \end{tabular}
        \label{tab:alpha}
        \vspace{-4mm}
\end{table}

\begin{table}[t]
    \renewcommand\arraystretch{0.9}
    \centering
    \caption{Results for different values of $\tilde{p}_{t}$ in FCBL.}
    \begin{tabular}{c|c|ccc}
    \toprule
    $\tilde{p}_{t}$ & $AP^b$  & $AP_r$  & $AP_c$  & $AP_f$  \\
    \midrule
    0.9 & 27.5 & 20.8 & 27.4 & 30.2 \\
    0.8 & 27.6 & 21.6 & 27.4 & 30.2 \\
    \textbf{0.7} & \textbf{27.8} & \textbf{20.1} & \textbf{28.2} & \textbf{30.3} \\
    0.6 & 27.6 & 21.0 & 27.6 & 30.2 \\
    0.5 & 27.5 & 20.9 & 27.5 & 30.1 \\
    \bottomrule
    \end{tabular}
    \label{tab:thres}
    \vspace{-4mm}
\end{table}

\begin{table}[t]
        \renewcommand\arraystretch{0.9}
        \centering
        \caption{Results for different values of $\beta$ in FHM.}
        \begin{tabular}{c|c|ccc}
        \toprule
        $\beta$         & $AP^b$  & $AP_r$  & $AP_c$  & $AP_f$           \\
        \midrule
        0.95 & 27.1 & 20.1 & 26.9 & 30.2 \\
        \textbf{0.9} & \textbf{27.8} & \textbf{20.1} & \textbf{28.2} & \textbf{30.3} \\
        0.85         & 27.7 &  20.2 & 27.9 & 30.3 \\
        \bottomrule
        \end{tabular}
        \label{tab:beta}
        \vspace{-4mm}
\end{table}

\begin{table}[t]
    \renewcommand\arraystretch{0.9}
    \centering
    \caption{Results for different values of $c,m$ in FHM.}
    \begin{tabular}{cc|c|ccc}
    \toprule
    $c$  & $m$ & $AP^b$   & $AP_r$  & $AP_c$  & $AP_f$  \\
    \midrule
    12 & 12 & 27.3 & 20.4 & 27.2 & 30.3 \\
    8 & 16 & 27.6 & 20.5 & 27.7 & 30.2 \\
    \textbf{8}  & \textbf{12} & \textbf{27.8} & \textbf{20.1} & \textbf{28.2} & \textbf{30.3} \\
    8  & 8 & 27.3 & 20.5 & 27.2 & 30.2 \\
    4  & 12 & 27.5 & 20.8 & 27.5 & 30.2 \\
    \bottomrule
    \end{tabular}
    \label{tab:km}
    \vspace{-4mm}
\end{table}

\begin{table}[t]
    \renewcommand\arraystretch{0.9}
    \centering
    \caption{Results for two formulas of the confusion matrix.}
    \begin{tabular}{c|c|ccc}
    \toprule
    $M_{i,j}$         & $AP^b$  & $AP_r$  & $AP_c$  & $AP_f$           \\
    \midrule
    \textbf{Eq in Last Row of Tab. \ref{tab:long-term indicator}} & \textbf{27.8} & \textbf{20.1} & \textbf{28.2} & \textbf{30.3} \\
    Eq. \eqref{eqn:19} & 27.3 & 19.9 & 27.2 & 30.3 \\
    \bottomrule
    \end{tabular}
    \label{tab:cm}
    \vspace{-4mm}
\end{table}

\begin{table}[t]
    \renewcommand\arraystretch{0.9}
    \centering
    \caption{Results for different formulations of the weight term.}
    \begin{tabular}{cc|c|ccc}
    \toprule
    $\tilde{p}_{j}\geq \tilde{p}_{i}$ & $\tilde{p}_{j}\geq \tilde{p}_{t}$   & $AP^b$  & $AP_r$  & $AP_c$  & $AP_f$           \\
    \midrule
    \checkmark &  & 27.5 & 20.8 & 27.5 & 30.2 \\
    & \checkmark & 20.1 & 11.1 & 20.5 & 23.0 \\
    \checkmark & \checkmark & \textbf{27.8} & \textbf{20.1} & \textbf{28.2} & \textbf{30.3} \\
    \bottomrule
    \end{tabular}
    \label{tab:wt}
    \vspace{-4mm}
\end{table}

\begin{table}[!t]
    \renewcommand\arraystretch{0.9}
    \centering
    \caption{Results for two kinds of training pipelines.}
    \begin{tabular}{c|c|ccc}
    \toprule
    Strategy & $AP^b$  & $AP_r$  & $AP_c$  & $AP_f$           \\
    \midrule
    End-to-end & 22.0 & 6.3 & 21.5 & 28.9 \\
    Decoupled & \textbf{27.8} & \textbf{20.1} & \textbf{28.2} & \textbf{30.3} \\
    \bottomrule
    \end{tabular}
    \label{tab:pipeline}
    \vspace{-4mm}
\end{table}

\begin{table}[!t]
    \renewcommand\arraystretch{0.9}
    \centering
    \caption{Results for FLOPs, Params, and FPS}
    \begin{tabular}{c|c|ccc}
    \toprule
    Backbone & Method & FLOPs (G) & Params (M) & FPS (img/s)           \\
    \midrule
    \multirow{2}{*}{R-50-FPN} & Baseline & 221.71 & 47.42 & 3.3 \\
    & BACL & 221.71 & 47.42 & 1.2 \\
    \midrule
    \multirow{2}{*}{R-101-FPN} & Baseline & 301.58 & 66.41 & 3.4 \\
    & BACL & 301.58 & 66.41 & 1.1 \\
    \bottomrule
    \end{tabular}
    \label{tab:speed}
    \vspace{-4mm}
\end{table}

\begin{table}[t]
    \centering
    \caption{Results for the baseline model and BACL on various architectures and backbones on LVIS v1.0 \texttt{val} set.}
    \resizebox{\columnwidth}{!}{%
    \begin{tabular}{c|c|c|c|ccc}
    \toprule
    Architecture                          & Backbone                        & BACL & $AP^b$           & $AP_r$           & $AP_c$           & $AP_f$           \\ \midrule
    \multirow{4}{*}{Faster R-CNN\cite{ren2015faster}}         & \multirow{2}{*}{R-50-FPN}  &  \ding{55}   & 19.3            & 1.1             & 16.1          & \textbf{30.9}          \\
                                          &                                 &   \ding{51}  & \textbf{26.1} & \textbf{16.0} & \textbf{25.6} & \textbf{30.9} \\ \cline{2-7} 
                                          & \multirow{2}{*}{R-101-FPN} &  \ding{55}   & 20.9          & 1.0           & 18.2          & \textbf{32.7} \\
                                          &                                 &   \ding{51}  & \textbf{27.8} & \textbf{18.1} & \textbf{27.3} & 32.6          \\ \midrule
    \multirow{4}{*}{Cascade Faster R-CNN\cite{cai2018cascade}} & \multirow{2}{*}{R-50-FPN}  &  \ding{55}   &      22.7       &      1.5         &      20.6         &     \textbf{34.4}          \\
                                          &                                 &   \ding{51}  &      \textbf{28.6}         &    \textbf{21.3}           &    \textbf{27.7}           &   32.8            \\ \cline{2-7} 
                                          & \multirow{2}{*}{R-101-FPN} & \ding{55}    &    24.5           &     2.6          &    23.1           &   \textbf{35.8}            \\
                                          &                                 &   \ding{51}  &       \textbf{29.8}        &      \textbf{22.0}         &     \textbf{28.8}          &    34.3           \\ \bottomrule
    \end{tabular}}
    \label{tab:benchmark}
\end{table}

\begin{figure}[t]
    \centering
    \includegraphics[width=\columnwidth]{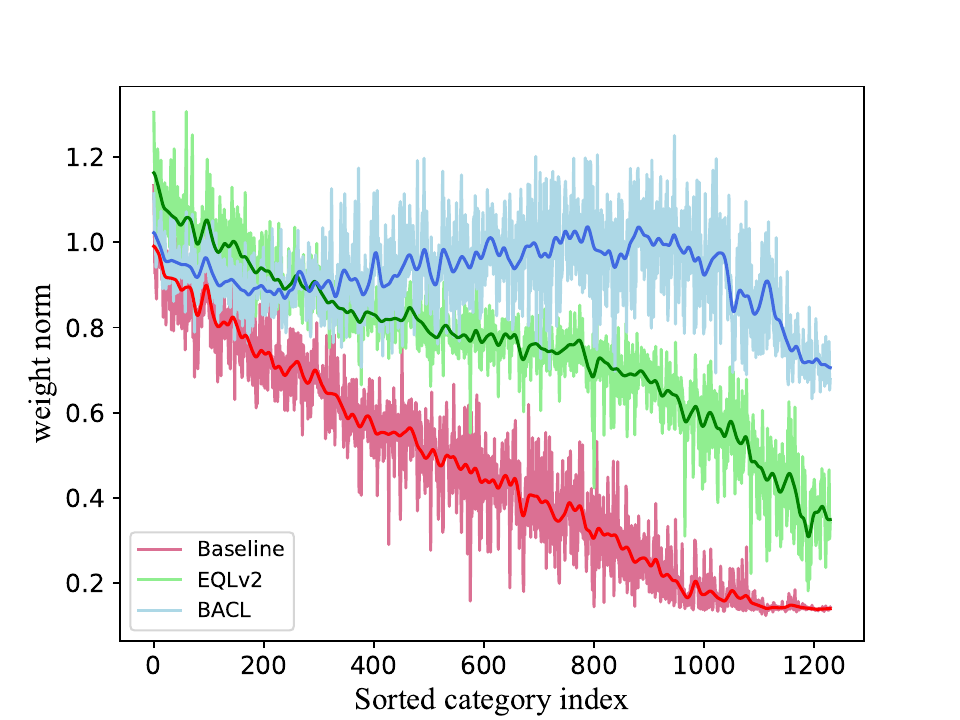}
    \caption{The L2 weight norm of the classifier of various models.}
    \label{pic:6}
    \vspace{-5mm}
\end{figure}

\subsection{Ablation Study}

We perform a series of ablation studies to better comprehend BACL.
All related experiments are conducted on LVIS v0.5 by using Faster R-CNN with ResNet-50-FPN backbone.
% In this section, we conduct a comprehensive ablation study to analyze the effects of various long-term indicators, different components and hyper-parameters in the classifier learning stage of BACL.
% All related experiments are conducted on the challenging LVIS v0.5 dataset by Faster R-CNN with ResNet-50-FPN backbone, which is fine-tuned on .

\noindent\textbf{Effectiveness of training strategies utilized in the representation learning stage.}
Tab. \ref{tab:representation_learning} shows the effects of the proposed training strategies in the representation learning stage.
The baseline model performs dismally on tail classes, achieving 1.6\% and 15.0\% AP for rare and common categories.
Sigmoid-based classifier marginally improves the detection performance for tail classes but leads to a 0.6\% decline in AP for head categories.
However, the inclusion of the objectness branch yields remarkable performance gains across all category groups, resulting in an overall AP of 20.5\%.
Besides, doubling the number of proposals kept after non-maximum suppression and adopting a small weight decay coefficient further enhances performance, particularly for $AP_r$ and $AP_c$.
With the simple Copy-Paste augmentation method and all the above strategies, the final detector achieves an AP of 21.7\% in the representation learning stage, outperforming the baseline model by a large margin.
These ablation experiments confirm the effectiveness of our utilized training strategies for representation learning.

\noindent\textbf{Comparison of various long-term indicators.}
In the classifier learning stage, we conduct a sequential evaluation of the long-term indicators listed in Tab. \ref{tab:long-term indicator} within BACL, and present the corresponding results in Fig. \ref{pic:5}.
The models are initialized using the checkpoint from the last row of Tab. \ref{tab:representation_learning}, which achieves an overall precision of 21.7\%.
Interestingly, both the one-dimensional static statistics and first-order dynamic statistics exhibit comparable performance in terms of overall precision (blue line).
This unexpected observation challenges our initial conjecture and earlier research, indicating that within a well-designed framework, these two types of statistics can have similar impacts.
Furthermore, the confusion matrix and true positive rate showcase the highest overall performances, ranking first and second, respectively.
Notably, when employed as a long-term indicator, the true positive rate achieves the highest precision for both rare (orange line) and frequent (yellow line) categories.
However, a substantial gap between the confusion matrix and other statistics in terms of $AP_c$ (gray line) establishes the superior overall performance of the former.
Therefore, we conclude that modeling inter-class relationships is advantageous for debiasing when dealing with long-tail distributions.
Thereinafter, all experiments adopt the confusion matrix as the long-term indicator.
% The statistics listed in Table \ref{tab:long-term indicator} are employed as the long-term indicators in turn and the experimental results are displayed in Fig. \ref{pic:5}.

\noindent\textbf{Component Analysis.}
As described in Section \ref{sec:3.4} and \ref{sec:3.5}, BACL integrates three components to calibrate classification bias in the classifier learning stage: an adaptive class-aware margin, an auto-adjusted weight term, and a dynamic feature hallucination module.
A series of experiments are conducted with different combinations of these components, and the corresponding results are recorded in Tab. \ref{tab:ca}.

The initial findings (first four lines) demonstrate the individual effectiveness of each component in improving the detector's performance on long-tailed data.
Specifically, the class-aware margin enhances $AP_r$ and $AP_c$ by 4.7\% and 3.5\% AP, respectively, without compromising the performance on frequent categories.
Similarly, the auto-adjusted weight term yields comparable outcomes.
However, the weight term slightly improves discrimination for head categories, while its impact on tail categories is less significant compared to the margin.
Combining them, FCBL elevates the performance of each category group and improves overall precision by 2.2\% AP, compared to the baseline model.
This improvement validates that FCBL effectively reduces misclassification between imbalanced foreground categories.

Furthermore, FHM achieves a 26.5\% AP with the ResNet-50-FPN backbone on LVIS v0.5, outperforming the baseline model by 4.5\% AP.
This result even surpasses most state-of-the-art methods, as indicated in Tab. \ref{tab:LVIS}.
Notably, this module, tailored for diversifying samples especially from tail classes, leads to a significant improvement in $AP_r$ (+13.9\% AP for rare categories) and $AP_c$ (+5.5\% AP for common categories).
More surprisingly, FHM obtains comparable performance to the baseline model for head classes, with 29.8\% AP versus 30.1\% AP for frequent categories.
This finding underscores the critical role of increasing sample diversity in resolving the long-tail problem.

Finally, by incorporating all three components, BACL achieves an impressive 27.8\% AP, outperforming any other combination.
With 20.1\% AP and 28.2\% AP on $AP_r$ and $AP_c$, BACL significantly advances the equilibrium across classes.
% \subsubsection{Component Analysis.} Our proposed framework BACL incorporates three components: $CCA$-based loss margin, $PPV$-based re-weighting mechanism and feature hallucination module. Table \ref{tab:ca} lists the detection results of each proposed component. The designed margin and re-weighting improves the baseline by 2.4, 3.4 AP for overall AP and 6.5, 10.5 AP for AP$_r$, respectively. The combination of them achieves 25.5 AP, outperforming the re-weighting mechanism by 0.1 AP. FHM increases the most for AP$_r$ by 14.2 AP, which indicates the effectiveness of adding variation to tail categories. BACL combines three components and gets further improvements to 26.9 AP.

\noindent\textbf{Hyper-parameters.}
We conducted a comprehensive set of experiments to examine the effects of hyperparameters in different components, namely $\alpha,\tilde{p}_{t},\beta,c,m$.
Tab. \ref{tab:alpha} presents the investigation into the impact of $\alpha$ on the class-aware margin, which controls the magnitude.
When $\alpha$ is set to 0, the adaptive margin is removed, resulting in the lowest overall performance of 27.2\% AP.
The optimal value of $\alpha$ is 0.85, where $AP^b$, $AP_c$, and $AP_f$ all reach the peaks.
Deviating from this optimal value, performance on rare categories improves further at the expense of a slight degradation in performance on common and frequent categories.
Therefore, we set $\alpha$ to 0.85 to strike the best trade-off.

In Tab. \ref{tab:thres}, we vary the probability threshold $\tilde{p}_t$ used in FCBL and find that the optimal results are obtained with a value of 0.7.
Straying from this threshold value leads to a degradation in overall performance.
Next, we investigate the effect of $\beta$, which determines the rate of updating feature distributions in FHM.
Tab. \ref{tab:beta} illustrates that assigning excessive weight to previous features significantly impacts the overall performance. 
Thus, we choose a value of 0.9 for better performance.
Regarding the number of sampled categories and synthesized hallucinated features, Tab. \ref{tab:km} indicates that the optimal settings are $c=8$ and $m=12$.

\noindent\textbf{Formulation of the confusion matrix.}
Similar to the advantages offered by soft labels compared to one-hot hard labels, the formulation for calculating the confusion matrix in Tab. \ref{tab:long-term indicator} is superior to its 
traditional definition:
\begin{equation}\label{eqn:19}
    M_{i,j} = \frac{\sum\limits_n \mathbb{I}[\mathop{\mathrm{argmax}}\limits_k p_{n,k}=j]\cdot\mathbb{I}[y_{n,i}=1]}{\sum\limits_n \mathbb{I}[y_{n,i}=1]},\; 1\le i,j\le C.
\end{equation}
The reason lies in the fact that the former considers the correlations between different classes as perceived by the classifier, whereas Eq. \eqref{eqn:19} only focuses on classification correctness.
As a result, the former provides more instructive information for calibration, as demonstrated in Tab. \ref{tab:cm}.

\noindent\textbf{Formulation of the weight term.}
In Eq. \eqref{eqn:14}, we compute the losses for misclassified categories as well for over-confident non-ground-truth categories.
To gain further insights, we conduct additional experiments by considering only one aspect at a time.
The results presented in Tab. \ref{tab:wt} shed light on the importance of the losses of misclassified classes for the detector.
When these losses are excluded, the model achieves only 20.1\% overall AP), despite a slight improvement in $AP_r$.
Besides, penalizing the over-confident non-ground-truth classes moderately improves performance by 0.3\% AP.

\noindent\textbf{Training pipelines.}
We evaluate BACL with end-to-end training pipelines, but the results are disappointing yielding only 22.0\% AP as shown in Tab. \ref{tab:pipeline}.
The reason might be that the unfrozen feature extractor will have a detrimental effect on the prototype of each category, preventing discriminatory hallucinated features from being synthesized by FHM.

% \subsubsection{Comparison of data augmentations used in the representation stage}
% As mentioned in section \ref{sec:3.2}, we provide proof

\noindent\textbf{Whether the classifier is balanced?}
As demonstrated in previous works~\cite{li2020overcoming}, the weight norm of a classifier is closely associated with the classification accuracy of its corresponding category.
Therefore, we visualize the weight norms of various classifiers trained by three methods: baseline, EQLv2, and BACL in Fig. \ref{pic:6}.
Apparently, the weight norm of the baseline model exhibits a positive correlation with the number of training instances in each category, thereby explaining the poor performance observed in tail categories.
While EQLv2 increases the weight norms for all classes, it still suffers from noticeable imbalances in weight norms
On the contrary, BACL has dramatically ameliorated the weight norm imbalance for each category and draws a nearly horizontal line, which interprets its balanced performance across all categories.

\noindent\textbf{FLOPs, Params and FPS.} We provide a comparison of floating point operations (FLOPs), parameters (Params), and frames per second (FPS) between the baseline model and BACL in Tab.~\ref{tab:speed}.
As BACL relies solely on classifier statistics for debiasing and does not introduce any trainable parameters, the number of parameters and FLOPs of BACL remain the same as those of the baseline model.
However, the FPS of the baseline model is nearly three times higher than that of BACL.
This can be explained based on the observation that the classifier supervised by BACL assigns higher scores to proposals, resulting in three times as many proposals being processed by the non-maximum suppression.
Since the non-maximum suppression runs on the CPU, the time required is proportional to the number of proposals being processed.

% Tables \ref{tab:alpha}-\ref{tab:km} shows that our method is not sensitive to every hyper-parameter in a wide range.
% The evaluation metric AP keeps relatively stable when these hyper-parameters have large changes.
% Without carefully searching for the optimal solution, we manually set $\alpha=0.75$ (for LVIS v0.5 only), $\beta=0.9,p_t=0.9,k=8,m=4$.

\subsection{Effectiveness of the Overall Framework}
Once we have identified the optimal long-term indicators and hyperparameters, we proceed to compare BACL with the baseline model, which is trained by the softmax cross-entropy loss, across various architectures and backbones. To validate the generalizability of BACL, we directly apply the optimal hyperparameters discovered on LVIS v0.5 to LVIS v1.0.

Tab. \ref{tab:benchmark} demonstrates that BACL consistently outperforms the baseline model by a large margin.
Examining the table reveals that Faster R-CNN with ResNet-50-FPN backbone exhibits an extremely imbalanced precision distribution under the supervision of the conventional softmax cross-entropy loss.
While it achieves satisfactory results on frequent categories with 30.9\% AP, the average precision on rare categories is nearly zero (1.1\% AP).
However, when trained using BACL, the same detector exhibits remarkable performance improvements of 14.9\% and 9.5\% AP on rare and common categories, respectively.
% Meanwhile, the performance of common classes has also been greatly improved, rising by 9.5\% AP.
The convincing improvements stem from the dual effect of alleviating unequal competition among foreground categories and enhancing sample diversities, especially in tail categories.
Moreover, the satisfactory performance on frequent categories remains intact, thanks in part to the adoption of advanced training strategies during the representation learning stage, which is the desired quality of a practical long-tail solution.
As a result, BACL yields an impressive 6.8\% increase in overall precision, elevating it from a discouraging 19.3\% AP to a promising 26.1\% AP.

Similar phenomena are observed when transitioning to a larger backbone network (ResNet-101-FPN) and a more intricate detector architecture (Cascade Faster R-CNN).
BACL consistently boosts performances on rare and common categories while maintaining satisfactory performance on frequent categories under these settings.
These results serve as compelling evidence of BACL's remarkable generalization ability.

\vspace{-2mm}
\subsection{Comparison with State-of-the-Arts}
\begin{table*}[t]
    \centering
    \caption{Performance comparison with state-of-the-art methods on LVIS {\em val} set. The ResNet-50-FPN and ResNet-101-FPN are adopted as backbones for Faster R-CNN. All methods are trained with a 2x schedule, {\em i.e.}, 24 epochs in total. BCE* denotes Sigmoid Cross-Entropy loss with an objectness branch. \textbf{Bold} numbers denote the best results.}
    \renewcommand\arraystretch{1.1}
    \resizebox{\textwidth}{!}{%
    \begin{tabular}{c|c|cccccccc|cccccccc}
        \hline
        \multirow{3}{*}{Strategy}    & \multirow{3}{*}{Methods} & \multicolumn{8}{c|}{LVIS   v0.5}                                                                                                                   & \multicolumn{8}{c}{LVIS   v1.0}                                                                                                                                                                \\ \cline{3-18} 
                                     &                          & \multicolumn{4}{c|}{ResNet-50-FPN}                                                 & \multicolumn{4}{c|}{ResNet-101-FPN}                           & \multicolumn{4}{c|}{ResNet-50-FPN}                                                                      & \multicolumn{4}{c}{ResNet-101-FPN}                                                 \\ \cline{3-18} 
                                     &                          & $AP_b$           & $AP_r$           & $AP_c$           & \multicolumn{1}{c|}{$AP_f$}           & $AP_b$           & $AP_r$           & $AP_c$           & $AP_f$           & \multicolumn{1}{c|}{$AP_b$}           & $AP_r$           & $AP_c$           & \multicolumn{1}{c|}{$AP_f$}           & \multicolumn{1}{c|}{$AP_b$}           & $AP_r$           & $AP_c$           & $AP_f$           \\ \hline\hline
        \multirow{14}{*}{End-to-end} & SCE\cite{ren2015faster}                     & 22.0          & 4.0           & 21.2          & \multicolumn{1}{c|}{30.1}          & 23.3          & 2.8           & 23.2          & 31.5          & \multicolumn{1}{c|}{19.3}          & 1.1           & 16.1          & \multicolumn{1}{c|}{30.9}          & \multicolumn{1}{c|}{20.9}          & 1.0           & 18.2          & 32.7          \\
                                     & BCE\cite{ren2015faster}               & 21.8          & 5.3           & 21.0          & \multicolumn{1}{c|}{29.5}          & 23.7          & 5.7           & 23.3          & 31.3          & \multicolumn{1}{c|}{19.5}          & 1.6           & 16.6          & \multicolumn{1}{c|}{30.6}          & \multicolumn{1}{c|}{21.4}          & 2.0           & 19.3          & 32.3          \\
                                     & BCE*\cite{ren2015faster}                   & 23.9          & 6.7           & 23.8          & \multicolumn{1}{c|}{31.0}          & 25.2          & 7.8           & 25.2          & 32.0          & \multicolumn{1}{c|}{21.6}          & 2.6           & 19.7          & \multicolumn{1}{c|}{32.1}          & \multicolumn{1}{c|}{23.1}          & 3.7           & 21.4          & \textbf{33.5} \\
                                     & RFS\cite{gupta2019lvis}                      & 25.0          & 14.1          & 24.8          & \multicolumn{1}{c|}{29.6}          & 25.9          & 14.8          & 25.5          & 30.8          & \multicolumn{1}{c|}{24.2}          & 14.2          & 22.3          & \multicolumn{1}{c|}{30.6}          & \multicolumn{1}{c|}{25.7}          & 15.9          & 23.7          & 32.2          \\
                                     & EQL\cite{tan2020equalization}                      & 24.0          & 9.4           & 24.4          & \multicolumn{1}{c|}{29.2}          & 25.5          & 9.9           & 26.1          & 31.1          & \multicolumn{1}{c|}{21.8}          & 3.6           & 21.1          & \multicolumn{1}{c|}{30.5}          & \multicolumn{1}{c|}{23.4}          & 4.5           & 22.9          & 32.3          \\
                                     & DropLoss\cite{hsieh2021droploss}                  & 23.3          & 9.7           & 24.7          & \multicolumn{1}{c|}{27.1}          & 26.1          & 11.2          & 28.5          & 29.0          & \multicolumn{1}{c|}{21.8}          & 5.2           & 21.8          & \multicolumn{1}{c|}{29.1}          & \multicolumn{1}{c|}{23.5}          & 5.9           & 23.9          & 30.7          \\
                                     & RIO\cite{chang2021image}                      & 24.4          & 15.7          & 24.0          & \multicolumn{1}{c|}{28.4}          & 26.4          & 16.4          & 26.4          & 30.5          & \multicolumn{1}{c|}{23.4}          & 15.3          & 21.2          & \multicolumn{1}{c|}{29.4}          & \multicolumn{1}{c|}{25.5}          & 17.2          & 23.7          & 31.2          \\
                                     & Forest R-CNN\cite{wu2020forest}             & 26.0          & 16.6          & 26.3          & \multicolumn{1}{c|}{29.4}          & 26.9          & 15.2          & 27.6          & 30.0          & \multicolumn{1}{c|}{-}             & -             & -             & \multicolumn{1}{c|}{-}             & \multicolumn{1}{c|}{-}             & -             & -             & -             \\
                                     & BALMS\cite{ren2020balanced}                    & 25.5          & 17.6          & 25.0          & \multicolumn{1}{c|}{29.3}          & 27.2          & 17.3          & 27.3          & 31.0          & \multicolumn{1}{c|}{24.1}          & 15.2          & 23.0          & \multicolumn{1}{c|}{29.4}          & \multicolumn{1}{c|}{26.9}          & \textbf{18.5} & 25.2          & 32.4          \\
                                     & De-confound-TDE\cite{tang2020long}           & 25.3          & 13.2          & 25.4          & \multicolumn{1}{c|}{30.0}          & 27.3          & 14.3          & 28.0          & 31.5          & \multicolumn{1}{c|}{23.7}          & 10.0          & 22.4          & \multicolumn{1}{c|}{31.2}          & \multicolumn{1}{c|}{-}             & -             & -             & -             \\
                                     & EQLv2\cite{tan2021equalization}                    & 26.5          & 17.2          & 26.2          & \multicolumn{1}{c|}{30.7}          & 27.4          & 18.3          & 26.7          & 31.8          & \multicolumn{1}{c|}{25.4}          & 15.8          & 23.5          & \multicolumn{1}{c|}{31.7}          & \multicolumn{1}{c|}{26.8}          & 17.1          & 24.9          & 33.1          \\
                                     & Seesaw\cite{wang2021seesaw}                   & 26.3          & 16.3          & 26.1          & \multicolumn{1}{c|}{30.6}          & 27.3          & 16.8          & 26.8          & 32.0          & \multicolumn{1}{c|}{24.8}          & 14.8          & 22.7          & \multicolumn{1}{c|}{31.6}          & \multicolumn{1}{c|}{26.6}          & 14.9          & 25.2          & 33.3          \\
                                     & FASA\cite{zang2021fasa}                    & 23.6          & 10.7          & 22.8          & \multicolumn{1}{c|}{29.6}          & 24.3          & 11.3          & 23.2          & 30.9          & \multicolumn{1}{c|}{21.5}          & 7.4           & 19.2          & \multicolumn{1}{c|}{30.2}          & \multicolumn{1}{c|}{22.9}             & 9.0             & 20.6             & 31.6             \\
                                     & PCB\cite{he2022relieving}                      & 23.9          & 9.1           & 22.9          & \multicolumn{1}{c|}{\textbf{31.2}} & 26.5          & 11.4          & 26.2          & \textbf{32.9} & \multicolumn{1}{c|}{23.0}          & 6.2           & 21.5          & \multicolumn{1}{c|}{\textbf{32.2}} & \multicolumn{1}{c|}{24.6}          & 8.0           & 23.1          & \textbf{33.5} \\ \hline\hline
        \multirow{6}{*}{Decoupled}   & SimCal\cite{wang2020devil}                   & 22.5          & 13.6          & 20.3          & \multicolumn{1}{c|}{29.0}          & -             & -             & -             & \textbf{-}    & \multicolumn{1}{c|}{-}             & -             & -             & \multicolumn{1}{c|}{-}             & \multicolumn{1}{c|}{-}             & -             & -             & -             \\
                                     & BAGS\cite{li2020overcoming}                     & 25.5          & 16.8          & 25.6          & \multicolumn{1}{c|}{28.8}          & 26.6          & 16.2          & 26.7          & 30.7          & \multicolumn{1}{c|}{23.7}          & 14.2          & 22.2          & \multicolumn{1}{c|}{29.6}          & \multicolumn{1}{c|}{25.4}          & 14.9          & 25.2          & 31.4          \\
                                     & ACSL\cite{wang2021adaptive}                     & 23.7          & 14.8          & 23.5          & \multicolumn{1}{c|}{27.5}          & 25.7          & 16.5          & 25.8          & 29.1          & \multicolumn{1}{c|}{22.2}          & 9.9           & 21.3          & \multicolumn{1}{c|}{28.5}          & \multicolumn{1}{c|}{23.7}          & 11.0          & 23.0          & 30.2          \\
                                     & DisAlign\cite{zhang2021distribution}                 & 25.2          & 14.1          & 25.2          & \multicolumn{1}{c|}{29.5}          & 27.4          & 15.9          & 27.8          & 31.5          & \multicolumn{1}{c|}{20.9}          & 3.9           & 20.4          & \multicolumn{1}{c|}{29.0}          & \multicolumn{1}{c|}{25.5}          & 13.3          & 24.5          & 32.0          \\
                                     & LOCE\cite{feng2021exploring}                     & 26.7          & 18.3          & 27.5          & \multicolumn{1}{c|}{28.9}          & 27.9          & 21.9          & 27.7          & 30.5          & \multicolumn{1}{c|}{25.1}          & 15.7          & 24.2          & \multicolumn{1}{c|}{30.1}          & \multicolumn{1}{c|}{26.7}          & 18.4          & 25.5          & 31.7          \\\cline{2-18}
                                     & BACL                     & \textbf{27.8} & \textbf{20.1} & \textbf{28.2} & \multicolumn{1}{c|}{30.3}          & \textbf{29.4} & \textbf{22.1} & \textbf{30.1} & 31.3          & \multicolumn{1}{c|}{\textbf{26.1}} & \textbf{16.0} & \textbf{25.7} & \multicolumn{1}{c|}{30.9}          & \multicolumn{1}{c|}{\textbf{27.8}} & 18.1          & \textbf{27.3} & 32.6          \\ \hline
        \end{tabular}}
    \label{tab:LVIS}
    \vspace{-4mm}
\end{table*}

On both versions of the LVIS dataset, we compare our proposed framework BACL to most state-of-the-art methods, which are categorized into end-to-end and decoupled training pipelines.
To ensure a fair comparison, we train all end-to-end methods for 24 epochs and all decoupled methods for 12+12 epochs using Faster R-CNN with a random sampler.
Following the convention~\cite{tan2021equalization,feng2021exploring}, we adopt ResNet-50-FPN and ResNet-101-FPN as backbones.

\noindent\textbf{Results for LVIS v0.5.}
As shown in Tab. \ref{tab:LVIS}, BACL surpasses all previous state-of-the-art methods by more than 1\% AP on ResNet-50-FPN backbone and achieves 27.8\%, which is already comparable to the performance of other state-of-the-art methods with ResNet-101-FPN backbone.
Furthermore, BACL demonstrates superior performance in detecting both rare and common classes, surpassing the second-place LOCE by 1.8\% and 0.7\% AP, respectively.

In contrast to end-to-end training pipelines, BACL achieves comparable or slightly inferior results on frequent categories due to the frozen feature extractor in the classifier learning stage, lagging behind PCB with the optimal results by merely 0.9\% AP.
However, BACL outperforms all end-to-end training methods in terms of rare categories by at least 2.5\% AP.
The optimal end-to-end training method, EQLv2, is still inferior to BACL on $AP_r$ and $AP_c$, respectively, by 2.9\% and 2.0\% AP.
Besides, BACL leaves seesaw loss further apart, which is second only to concurrent EQLv2.
This outcome is attributed to the intensified sample diversity introduced by BACL.
Interestingly, FASA, which also emphasizes increasing data diversity, falls significantly short with a 23.6\% AP, making it nearly 3\% AP less effective than our proposed FHM, as shown in Tab. \ref{tab:ca}.
We attribute this outcome to the decoupled training pipeline and more accurate feature distributions.

We compare BACL with other decoupled training methods and observe that decoupled training methods have the advantage of improving the performance of rare classes, with BACL achieving over 20\% AP in this regard.
Notably, BACL manages to overcome the drawback of decoupled training methods, which is a decline in the performance of frequent categories, while achieving large improvements in $AP_r$ and $AP_c$.
As a result, BACL outperforms SimCal, BAGS, ACSL, DisAlign, and LOCE by a substantial margin.

When switching to a large backbone ResNet-101-FPN, BACL demonstrates improved overall performance with an AP of 29.4\%.
Moreover, there is a large enhancement of 1\% to 2\% AP in the detection results for each category group.

\noindent\textbf{Results for LVIS v1.0.}
Without performing an extensive search for optimal hyperparameters, we directly train the detectors on LVIS v1.0 \texttt{train} set with the BACL settings from LVIS v0.5.
As shown in Tab. \ref{tab:LVIS}, BACL remains at the forefront of the state-of-the-art methods, winning by at least 0.7\% AP with ResNet-50-FPN and 0.9\% AP with ResNet-101-FPN.
While BACL's superiority in rare classes may have diminished, it still outperforms other methods by at least 0.2\% AP with ResNet-50-FPN.
As for common classes, BACL surpasses the second-place LOCE by 1.5\% AP and 1.8\% AP on two backbones.
Besides, BACL achieves performance on par with the conventional cross-entropy loss in terms of $AP_f$, further reinforcing its practicality in real-world scenarios.

\begin{table}[!t]
    \centering
    \caption{Comparison results for instance segmentation on LVIS v1.0 \texttt{val} set. $AP^m, AP_r, AP_c, AP_f$ are mask APs.}
    \begin{tabular}{c|c|c|c|ccc}
        \toprule
        Backbone                        & BACL & $AP^m$   & $AP^b$   & $AP_r$   & $AP_c$   & $AP_f$   \\ \midrule
        \multirow{2}{*}{R-50-FPN}  & \ding{55}    & 19.4  & 19.9  & 1.3   & 17.2  & \textbf{29.9}  \\
                                        & \ding{51}    & \textbf{25.4} & \textbf{26.1} & \textbf{17.7} & \textbf{25.0} & 29.1  \\ \midrule
        \multirow{2}{*}{R-101-FPN} & \ding{55}    &   20.8    &    21.8   &   1.5    &    19.5   &  30.8     \\
                                        & \ding{51}    & \textbf{27.2} & \textbf{28.4} & \textbf{19.3} & \textbf{27.0} & \textbf{30.9} \\ \bottomrule
        \end{tabular}
    \label{tab:instance}
    \vspace{-4mm}
\end{table}
% BACL consistently outperforms the baseline model by 5.7 AP. Compared to state-of-the-art methods, BACL achieves comparable results. As shown in Table \ref{tab:LVIS v1}, end-to-end methods are a little superior to decoupled training methods. We attribute this phenomenon to sub-optimal feature extractor. The decoupled traing pipeline limits feature extractor to get further improvement, especially on larger dataset. The results for BACL* verify our assumption, which outperforms all state-of-the-art by at least 0.2 AP.
% \noindent\textbf{Stronger backbone.} BACL and BACL* consistently outperform the baseline and SOTA with stronger backbone, {\em i.e.}, ResNet-101-FPN, which shows high classification equilibrium.

\subsection{Extension to Long-Tailed Instance Segmentation}
We further evaluate BACL on the long-tailed instance segmentation task with Mask R-CNN.
The experimental results are summarized in Tab. \ref{tab:instance}.
BACL surpasses the baseline model by a large margin, with over 16\% AP improvement of $AP_r$ on both backbones.
Notably, it also boosts $AP_c$ and even achieves comparable performances with the baseline on $AP_f$.

\subsection{Results on COCO-LT}
To further verify the effectiveness of BACL, we conduct experiments on the COCO-LT dataset.
Following~\cite{wang2020devil}, we divide 80 categories into 4 groups with $<20$, 20-400, 400-8000, and $\ge 8000$ training instances and report the accuracy for each group as $AP_1$, $AP_2$, $AP_3$, $AP_4$.
In Tab. \ref{tab:cocolt}, we compare BACL with the baseline model and two leading methods.
The results demonstrate that BACL consistently outperforms the baseline model by over 2.0\% AP on overall accuracy, with a significant improvement of over 14\% AP on tail categories (groups 1 and 2).
When compared to Seesaw and EQLv2, our framework retains its dominant position, leading by over 1.1\% AP.
Remarkably, BACL even achieves substantial progress on the performance of tail categories in comparison with these two state-of-the-art methods on the LVIS dataset.
We conjecture that the reason lies in the fact that these methods only consider the problem of unequal competition between foreground categories without alleviating the problem of insufficient sample diversity in tail categories.
Instead, our framework tackles both issues through a divide-and-conquer approach.
 
\begin{figure}[t]
    \centering
    \subfigure{
        \includegraphics[width=2.1cm]{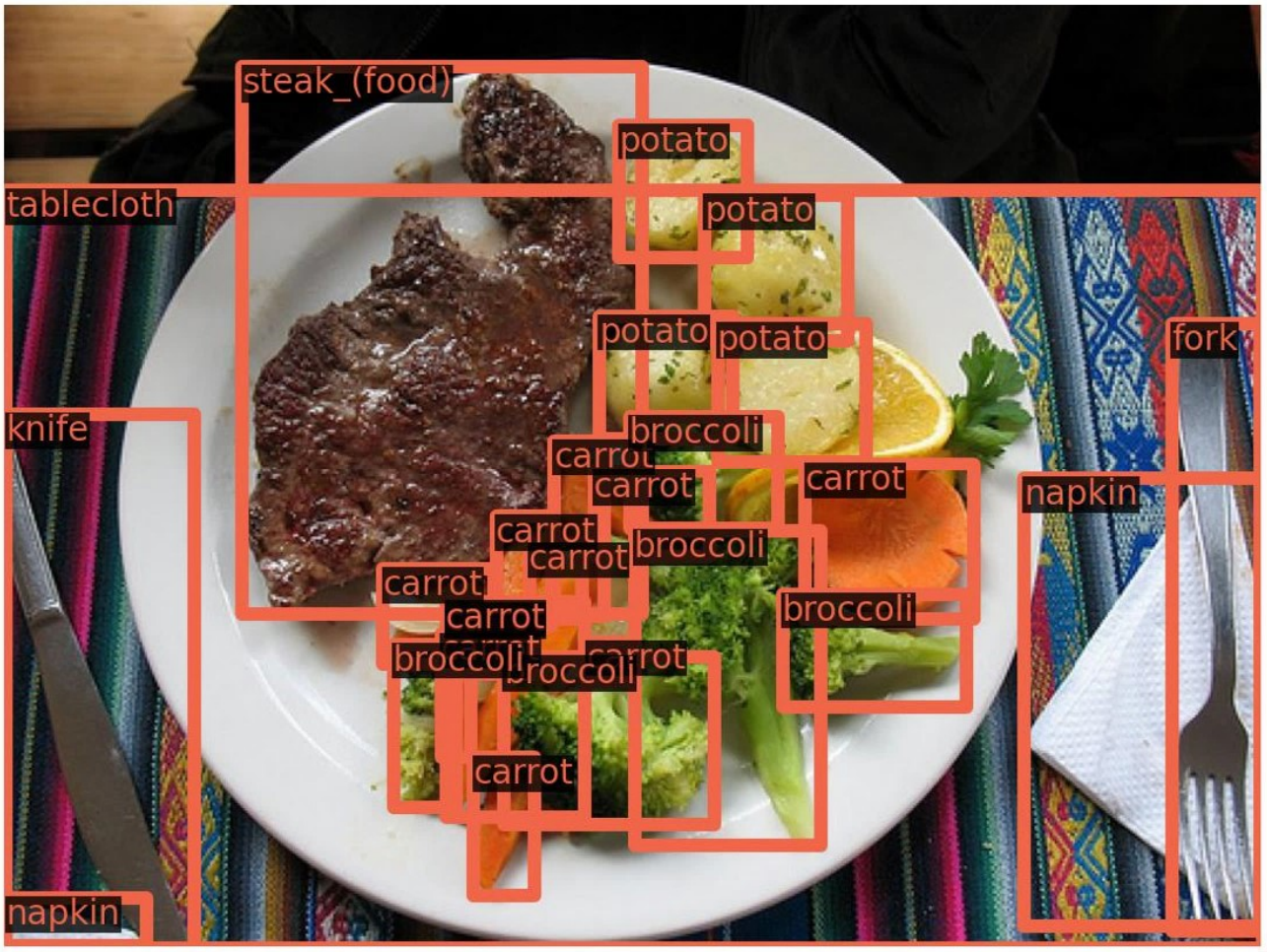}
    }\hspace{-3mm}
    \subfigure{
	\includegraphics[width=2.1cm]{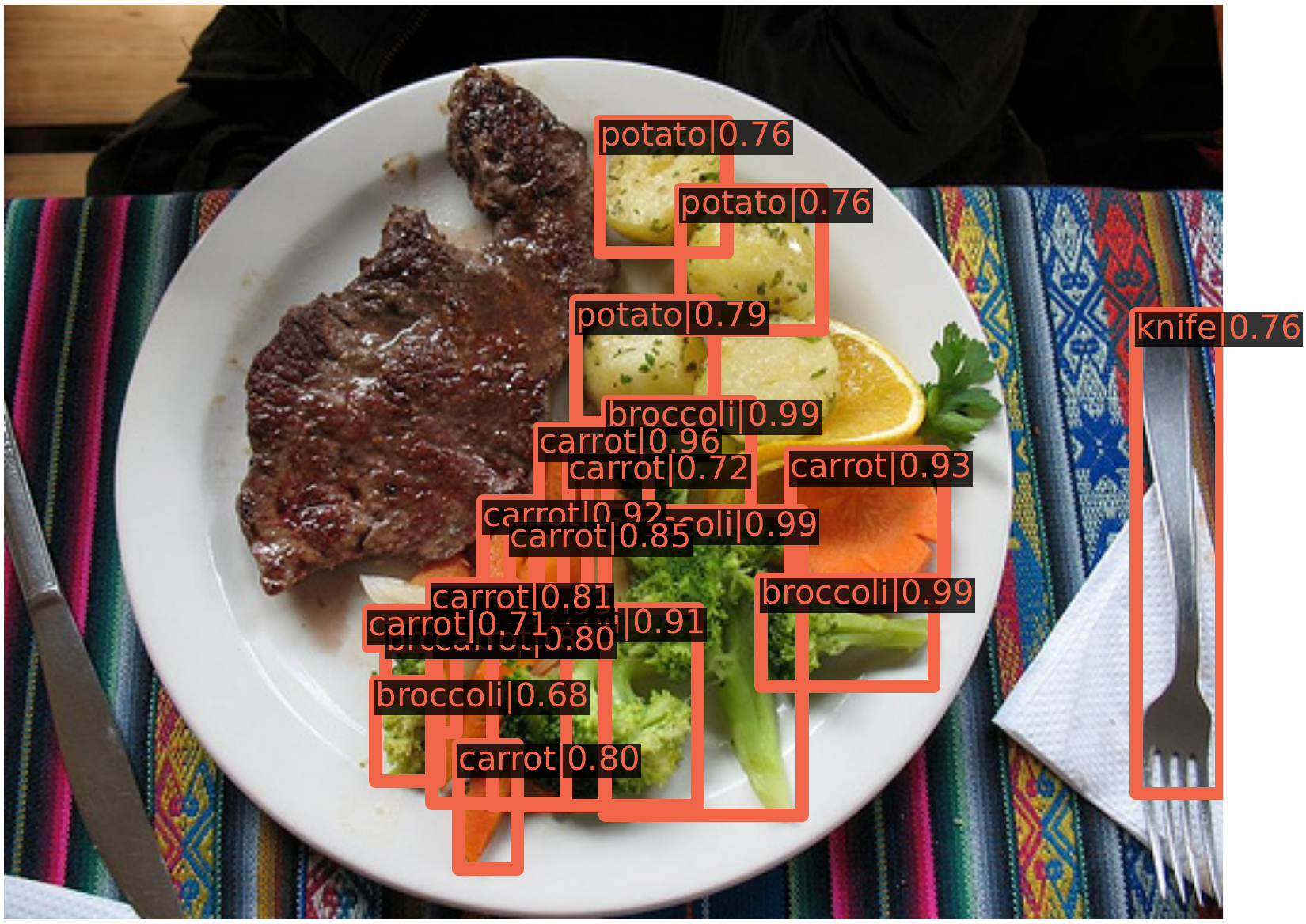}
    }\hspace{-3mm}
    \subfigure{
    	\includegraphics[width=2.1cm]{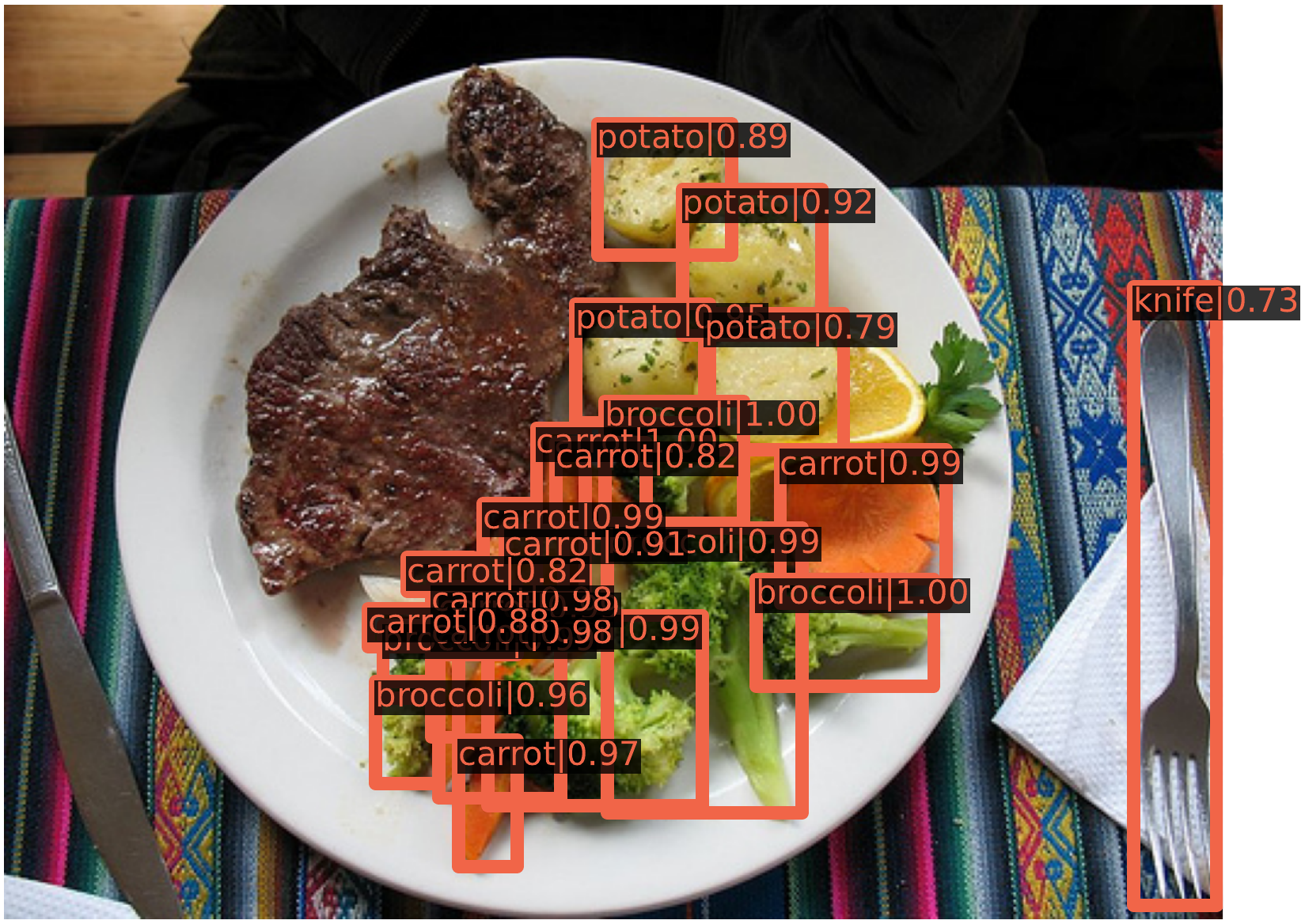}
    }\hspace{-3mm}
    \subfigure{
	\includegraphics[width=2.1cm]{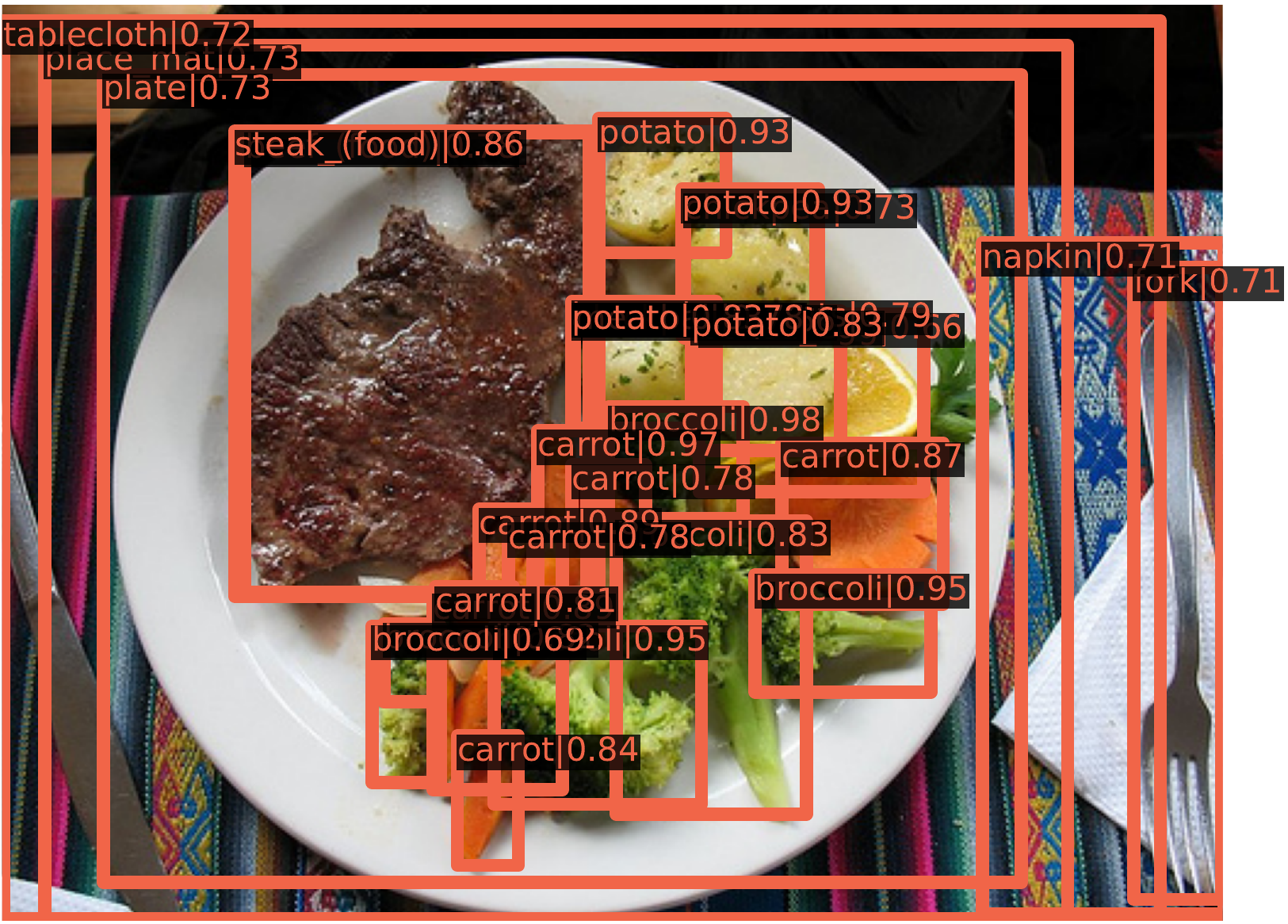}
    }\vspace{-1mm}
    \quad
    \subfigure{
        \includegraphics[width=2.1cm]{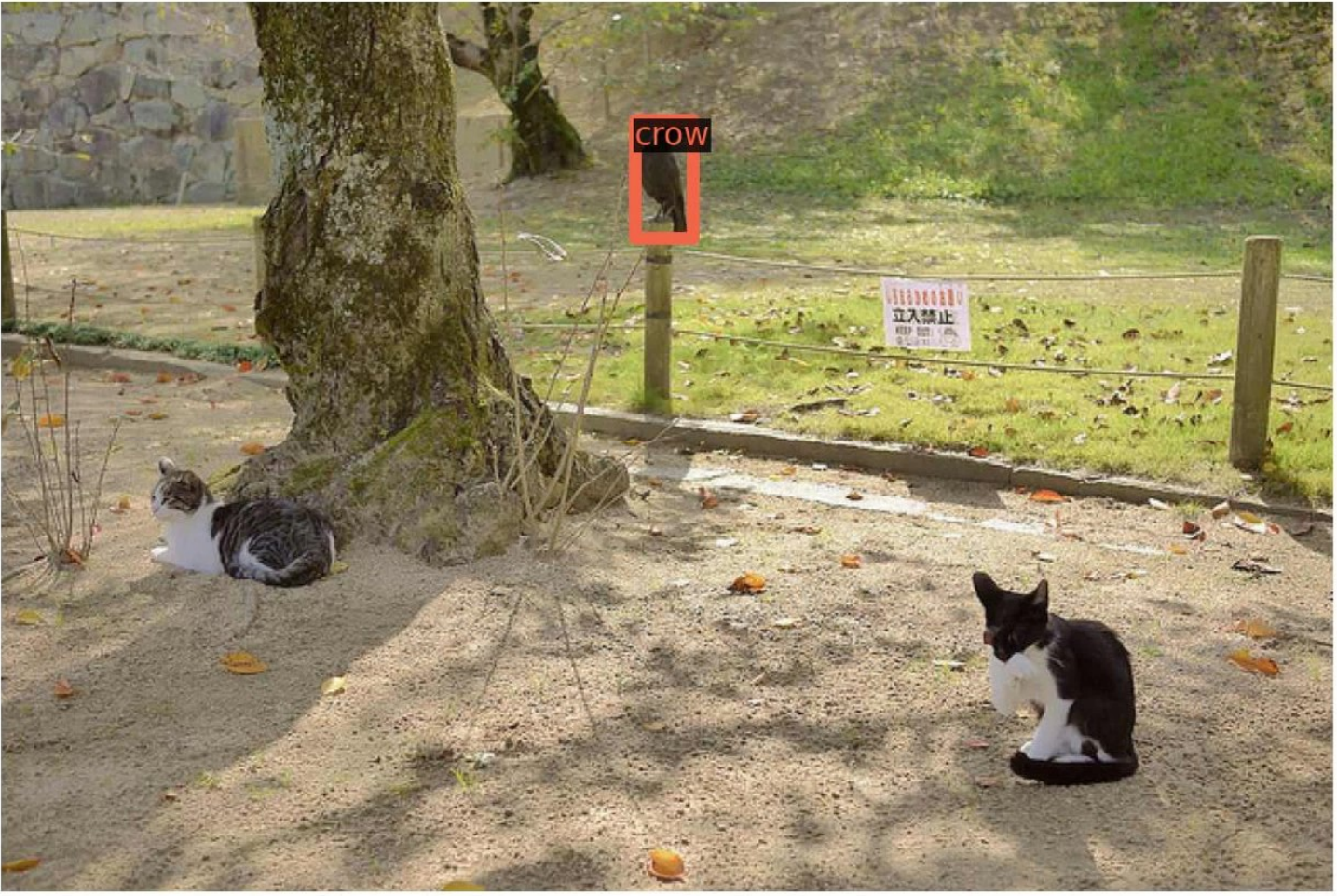}
    }\hspace{-3mm}
    \subfigure{
	\includegraphics[width=2.1cm]{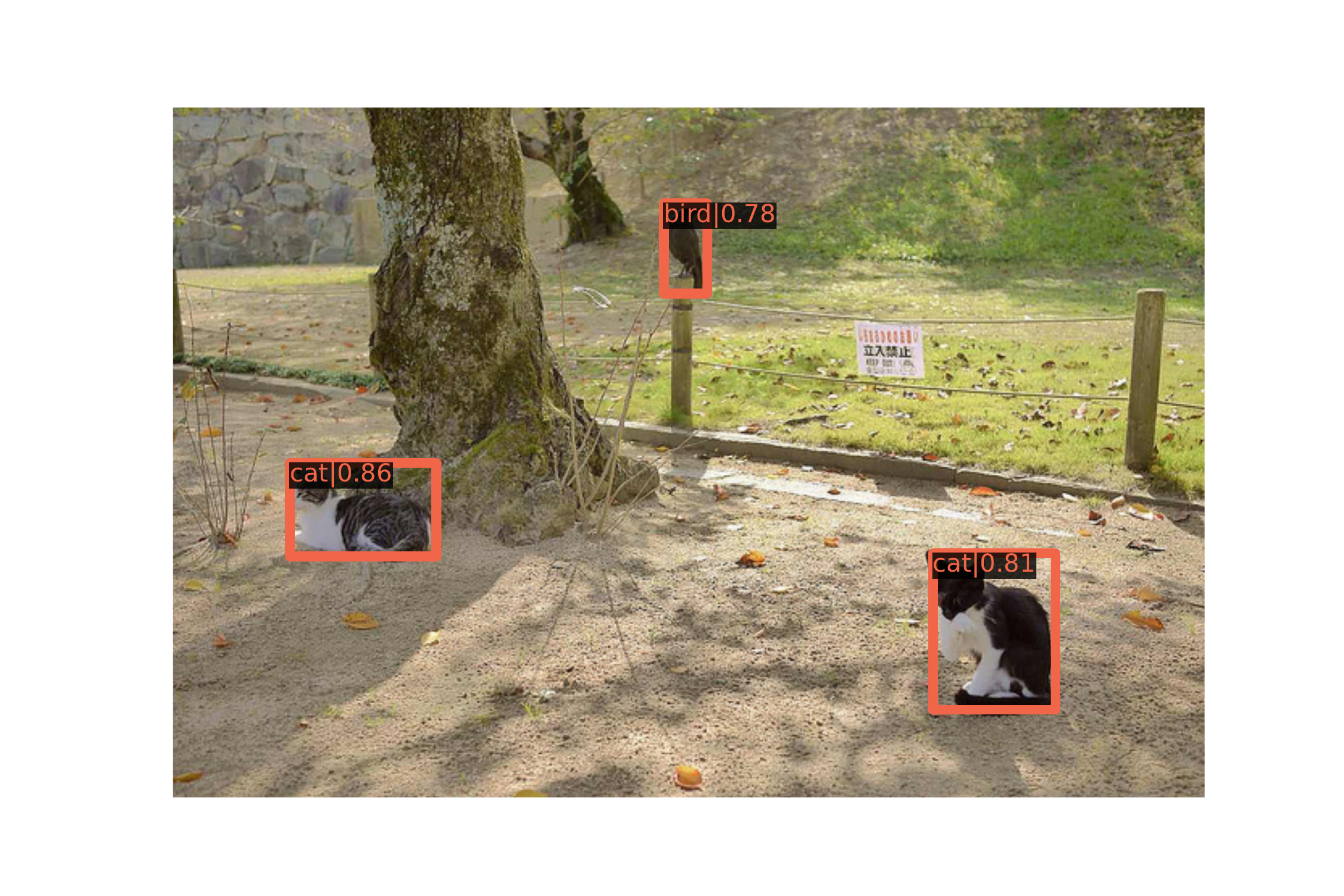}
    }\hspace{-3mm}
    \subfigure{
    	\includegraphics[width=2.1cm]{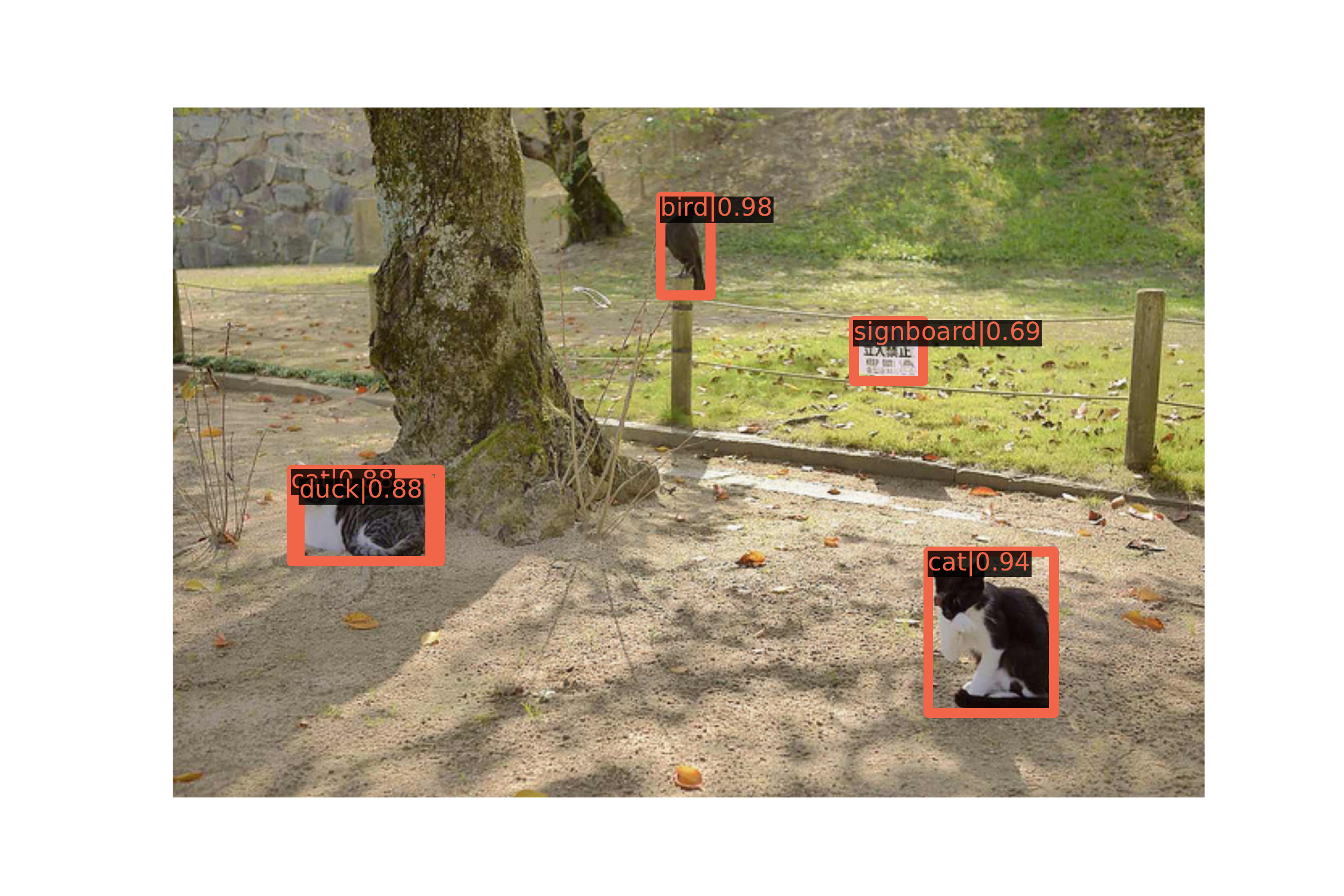}
    }\hspace{-3mm}
    \subfigure{
	\includegraphics[width=2.1cm]{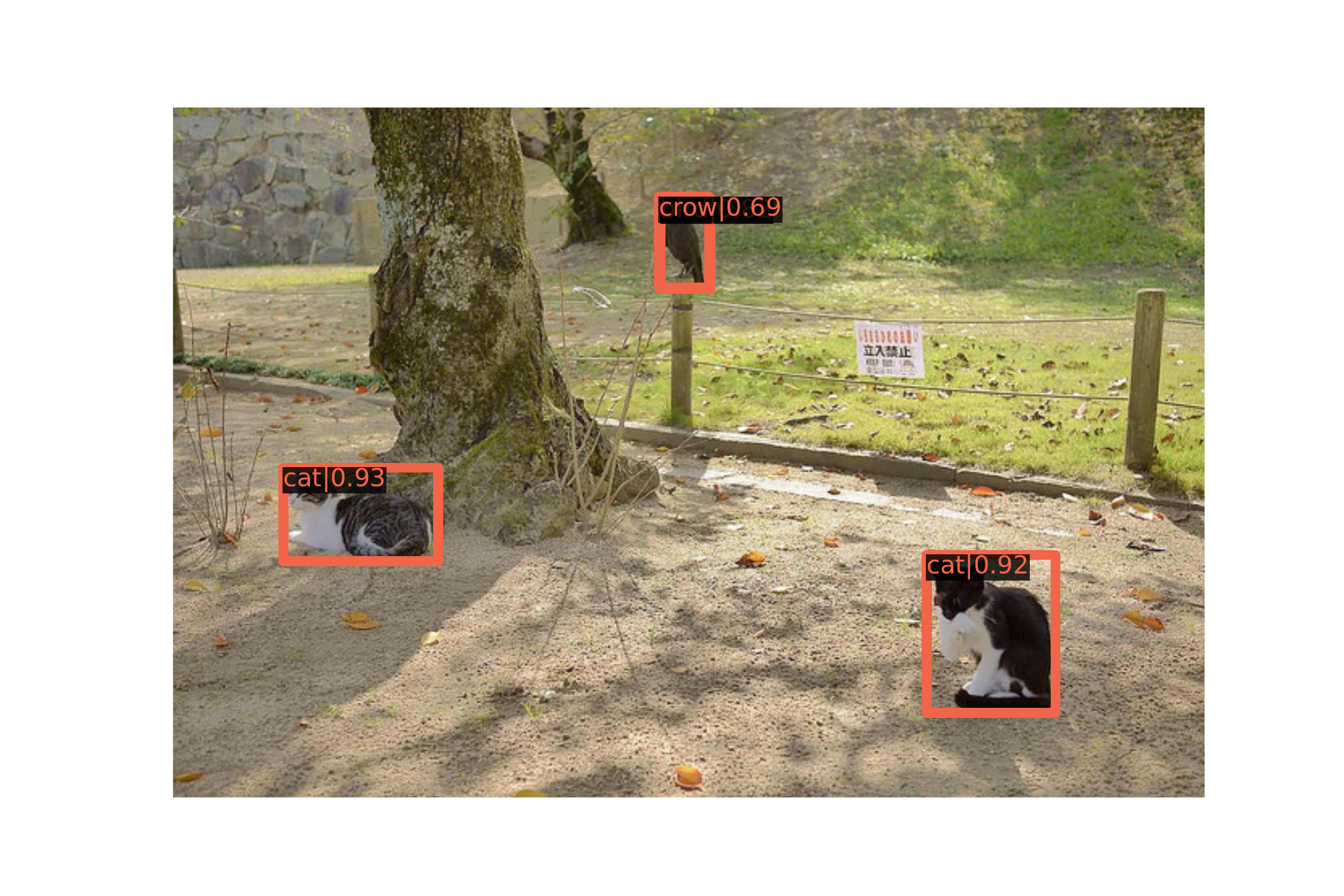}
    }\vspace{-1mm}
    \quad
    \setcounter{subfigure}{0}
    \subfigure[Ground Truth]{
	\includegraphics[width=2.1cm]{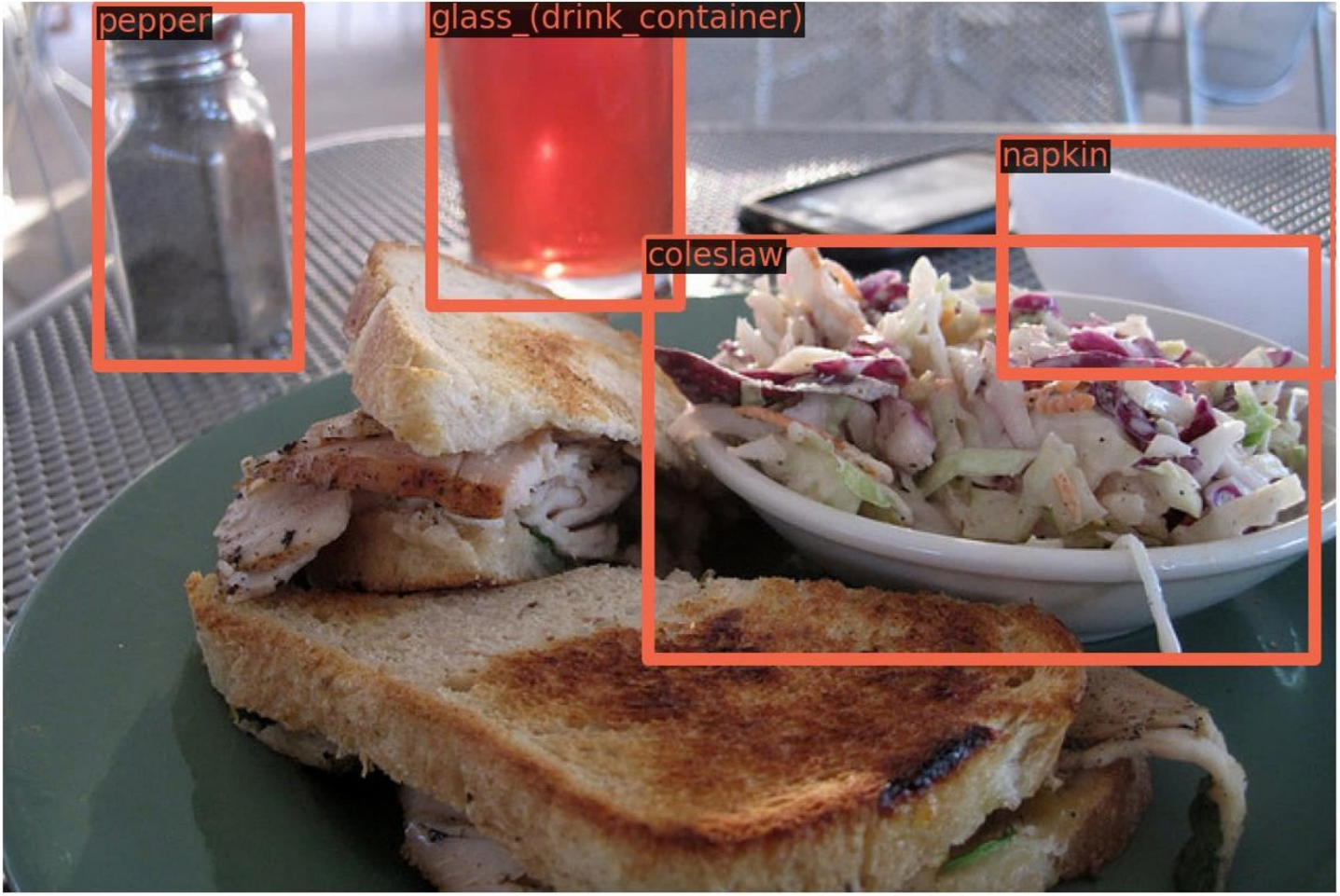}
        \label{gt}
    }\hspace{-3mm}
    \subfigure[Baseline]{
	\includegraphics[width=2.1cm]{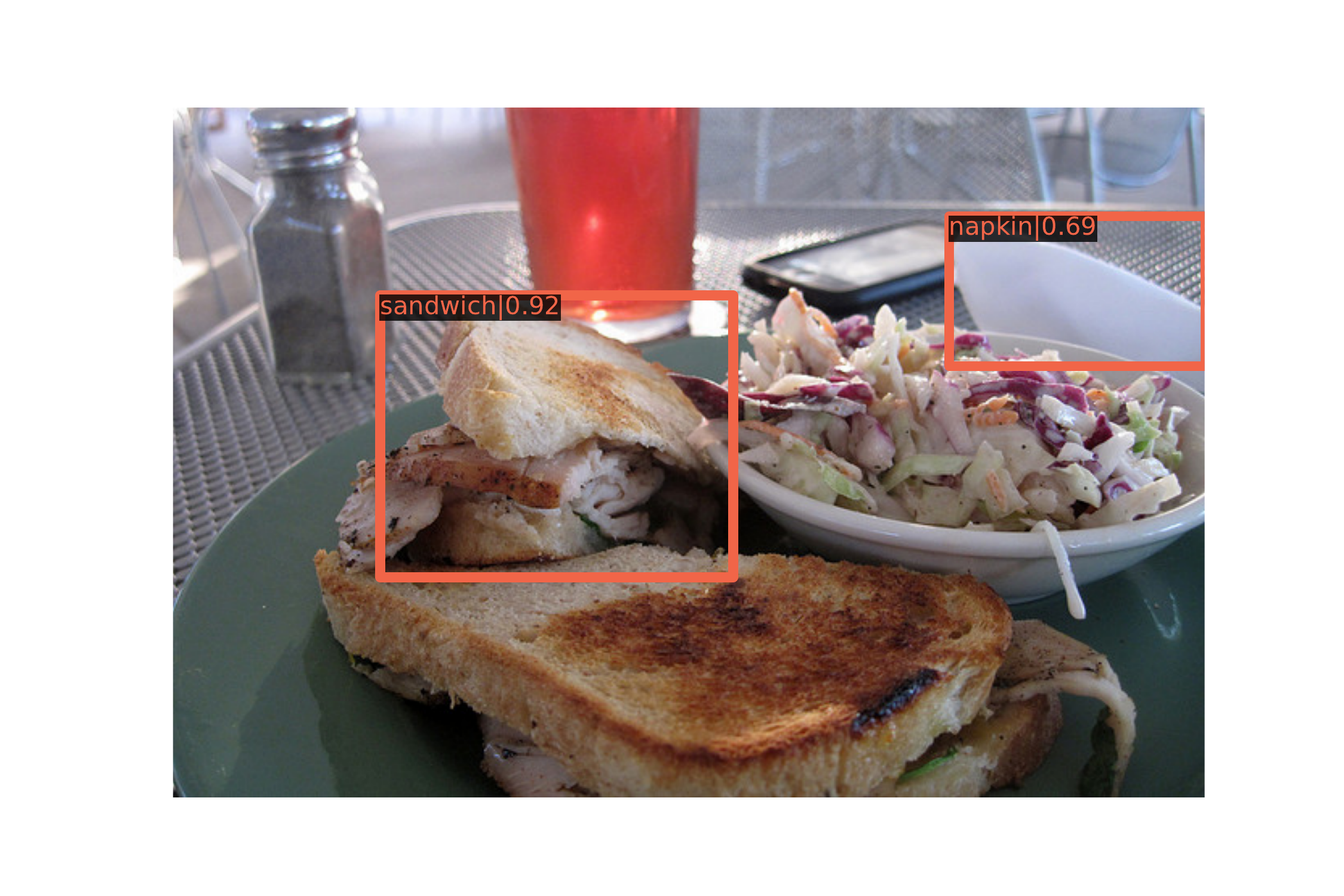}
        \label{baseline}
    }\hspace{-3mm}
    \subfigure[EQLv2]{
	\includegraphics[width=2.1cm]{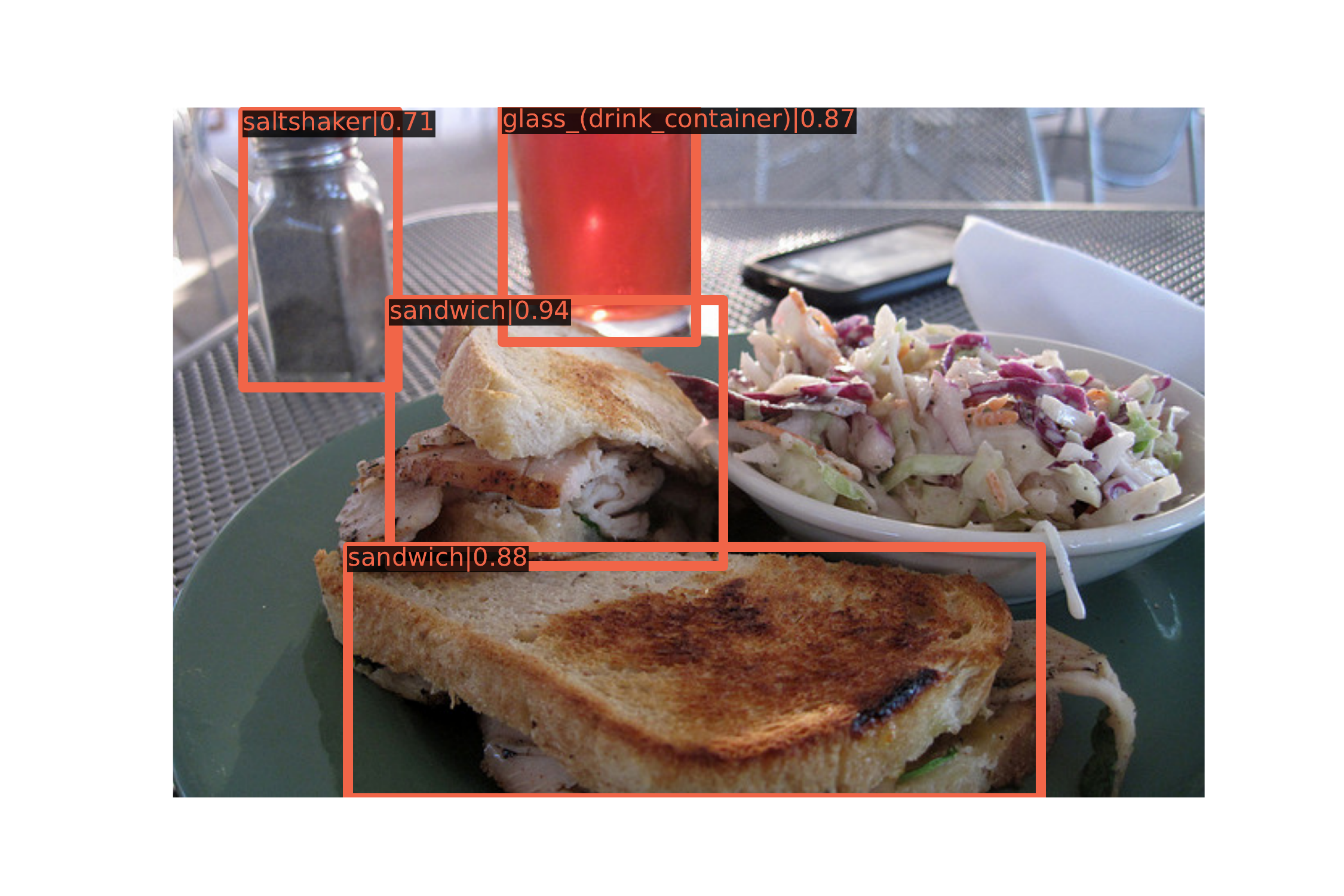}
        \label{eqlv2}
    }\hspace{-3mm}
    \subfigure[BACL]{
	\includegraphics[width=2.1cm]{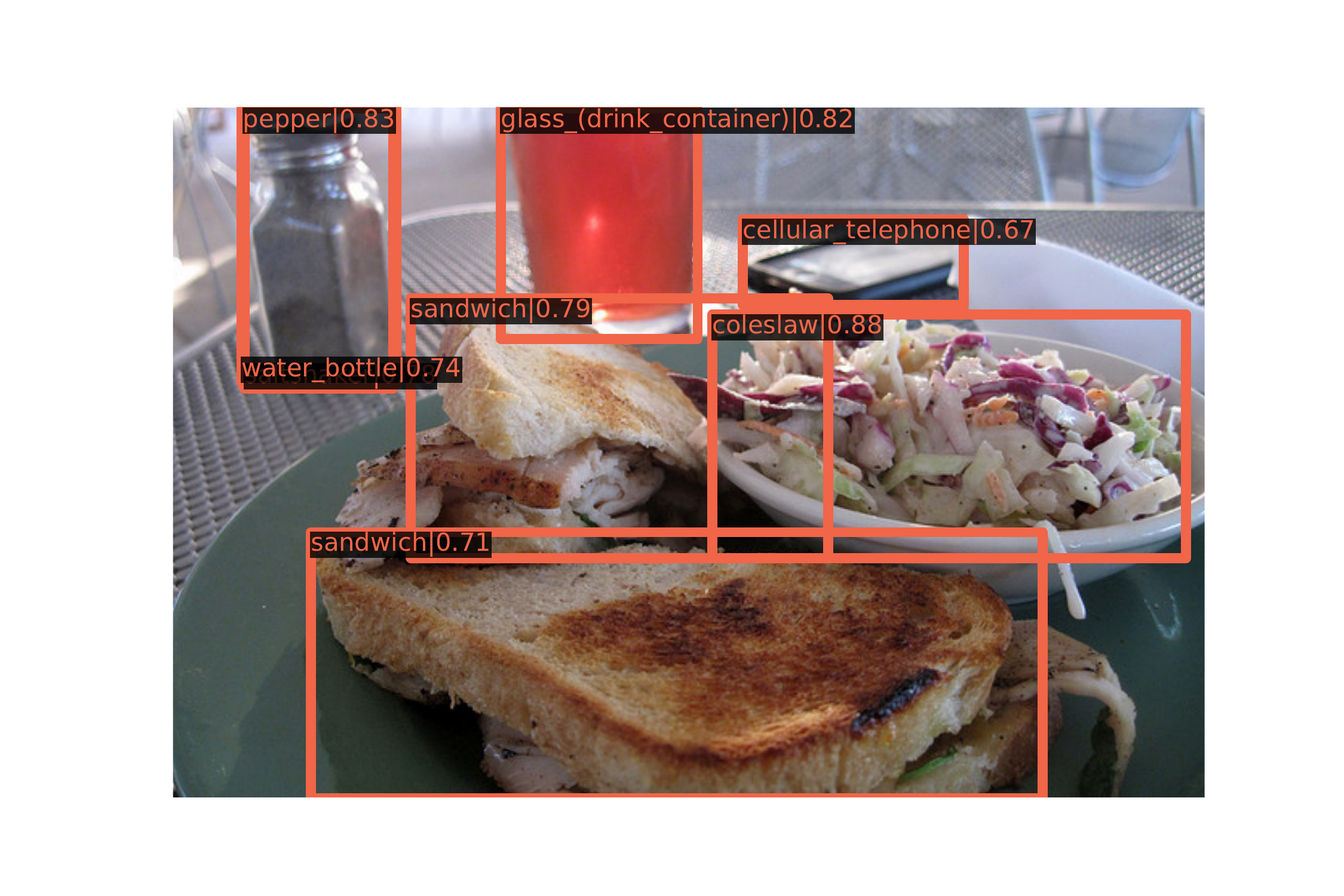}
        \label{bacl}
    }
    \caption{Several qualitative results on LVIS v0.5 \texttt{val} set.}
    \label{pic:7}
    \vspace{-2mm}
  \end{figure}

\begin{table}[t]
    \centering
    \caption{Results on COCO-LT \texttt{val} set.}
    \begin{tabular}{c|c|c|cccc}
        \toprule
        Backbone                        & Methods & $AP^b$   & $AP_1$   & $AP_2$   & $AP_3$   & $AP_4$   \\ \midrule
        \multirow{4}{*}{R-50-FPN}  & Baseline    & 24.5 & 0 & 14.6 & 29.6 & 32.9  \\
                                        & EQLv2    & 25.7 & 3.8 & 18.1 & 29.6 & \textbf{33.0}  \\
                                        & Seesaw    & 23.9 & 3.0 & 14.5 & 28.4 & 32.3  \\
                                        & BACL    & \textbf{26.8} & \textbf{14.4} & \textbf{20.8} & \textbf{29.7} & 31.9  \\ \midrule
        \multirow{4}{*}{R-101-FPN} & Baseline    &   26.0    &    0   &   16.4    &    \textbf{31.4}   &  \textbf{34.2}     \\
                                        & EQLv2    & 26.8 & 3.2 & 19.4 & 30.8 & 34.1  \\
                                        & Seesaw    & 24.9 & 3.2 & 14.5 & 30.0 & 33.4  \\
                                        & BACL    & \textbf{28.0} & \textbf{15.4} & \textbf{21.9} & 31.2 & 33.0  \\ \bottomrule
        \end{tabular}
    \label{tab:cocolt}
    \vspace{-3mm}
\end{table}

\subsection{Qualitative Results}
We provide several qualitative results of the baseline model, EQLv2, and BACL for comparison in Fig. \ref{pic:7}.
In the first image, only BACL accurately classifies the object as the frequent category "fork", whereas both the baseline and EQLv2 mistake it as "knife".
BACL even identifies the missing "steak" that other methods fail to detect.
Besides, BACL successfully categorizes the object as the rare category "crow" in the second row.
It also recognizes missing objects for common categories, such as "steak" in the first image, as well as "pepper" and "coleslaw" in the third image.
These observations demonstrate that the separation processing for foreground and background in BACL mitigates the issue of unequal competition among foreground categories while maintaining discrimination between foreground and background.
Moreover, the enhanced sample diversity in BACL contributes to the recognition of tail categories.

\vspace{-2mm}
\section{Conclusion}
This paper presents BACL, a unified framework for the long-tailed object detection task.
By adopting a divide-and-conquer approach, BACL introduces the FCBL to alleviate the unequal competition among foreground categories and the FHM to intensify the diversities of tail categories.
Extensive experiments demonstrate that BACL provides detectors with a more balanced and accurate classification branch across various backbones and architectures.
However, BACL is designed on the foundation of the decoupled training pipeline, restricting the improvement of the feature extractor in the classifier learning stage.
Future work may compensate for this drawback and integrate more advanced indicators and approaches to add sample variances for further improvement.
We believe that the proposed BACL can also be applied to other long-tailed recognition tasks after task-specific modifications.

\noindent\textbf{Acknowledgement:} We acknowledge the support of GPU cluster built by MCC Lab of Information Science and Technology Institution, USTC.
% In this paper, we propose a unified framework named BACL for long-tailed object detection.
% BACL creatively utilizes a long-short-term indicators pair to reflect the learning status of the classifier.
% Based on this pair, a Foreground Classification Balance Loss adaptively rectifies category distribution differences and focuses on confounding classes to boost classification equilibrium and correctness.
% Meanwhile, a Feature Hallucination Module synthesizes hallucinated RoI features to intensify sample diversities for tail classes.
% Extensive experiments demonstrate that BACL provides detectors with a more balanced and accurate classification branch across various backbones and architectures.
% In the future, we believe BACL could easily integrate advanced indicators and approaches to add sample variances for further improvement.

\vspace{-3mm}
\bibliographystyle{IEEEtran}
\bibliography{paper}

\vfill

\end{document}